\documentclass[11pt]{article}
\usepackage[american]{babel}
\usepackage[utf8]{inputenc} 
\usepackage{csquotes}
\usepackage[style=apa,sortcites=true,sorting=nyt,backend=biber]{biblatex}
\usepackage[margin=1in]{geometry}
\usepackage{setspace}
\usepackage{float}
\usepackage{hyperref}       
\usepackage{url}            
\usepackage{booktabs}       
\usepackage{amsfonts}       
\usepackage{pifont}
\usepackage{nicefrac}       
\usepackage{microtype}      
\usepackage{lipsum}
\usepackage{graphicx}
\usepackage{amsmath}
\usepackage{comment}
\usepackage{MnSymbol}
\usepackage{wasysym}
\usepackage[inkscapearea=page]{svg}
\usepackage{adjustbox}
\usepackage{booktabs} 
\usepackage{xparse}   
\usepackage[switch]{lineno}
\usepackage{caption}
\usepackage{subcaption}
\usepackage{array}
\usepackage{multirow}
\usepackage{placeins}
\usepackage{makecell}

\title{Generalization of Fine-Tuned Uncertainty Communication and Metacognition in Large Language Models}

\usepackage{authblk}
\author[1,2]{Mark Steyvers\thanks{Correspondence should be sent to \href{mailto:mark.steyvers@uci.edu}{mark.steyvers@uci.edu}}}
\author[2]{Catarina Belem}
\author[2]{Padhraic Smyth}

\affil[1]{Department of Cognitive Sciences, University of California, Irvine, United States}
\affil[2]{Department of Computer Science, University of California, Irvine, United States}

\date{}

\newif\ifcomments
\commentstrue
\ifcomments
\providecommand{\mark}[2][]{{\protect\color{teal}{[Mark:\textbf{#1} #2]}}}
\providecommand{\ps}[2][]{{\protect\color{green}{[Padhraic:\textbf{#1} #2]}}}
\providecommand{\kat}[2][]{{\protect\color{purple}{[Kat:\textbf{#1} #2]}}}
\else
\providecommand{\mark}[2][]{}
\providecommand{\ps}[2][]{}
\providecommand{\kat}[2][]{}
\fi

\begin{document}
\maketitle
 
\begin{abstract}
\textbf{Background}. Large language models are increasingly used in settings where confident but incorrect answers can mislead users. Reliable uncertainty communication requires a form of metacognition: monitoring when one’s own answers are likely to be correct. Yet models’ stated confidence is often poorly aligned with answer correctness. We test whether supervised fine-tuning improves uncertainty communication and whether gains transfer across domains and task formats.

\textbf{Methods}. We fine-tuned two models on general knowledge, mathematics, and open-ended trivia questions. We evaluated single-question confidence estimation, in which the model reports numeric confidence for one answer, and pairwise confidence comparison, in which it chooses which of two questions it is more likely to answer correctly. We tested held-out questions from training domains and new medical, legal, and truthfulness benchmarks. We assessed calibration, discrimination, and answer accuracy before and after fine-tuning.

\textbf{Results}. Here we show that fine-tuning improves alignment between stated confidence and observed accuracy and increases the model’s ability to assign higher confidence to correct than to incorrect answers. Gains occur within training domains and, to a lesser extent, in new domains. However, single-task training does not reliably transfer between single-question confidence estimation and pairwise confidence comparison. Multitask fine-tuning produces broader gains in the models and tasks studied here.

\textbf{Conclusions}. Uncertainty communication in large language models is trainable, but transfer across metacognitive tasks is limited. Joint training on multiple confidence tasks may support broader generalization, although further tests across model families and metacognitive tasks are needed.
\end{abstract}

\section*{Introduction}
What would it take for a language model not only to generate an answer but also to assess how likely that answer is to be correct? As Large Language Models (LLMs) are increasingly deployed in education, business, law, and medicine, their responses shape decisions with real-world consequences. However, these models sometimes produce factually incorrect information, and users may act on such errors when they are presented without clear signals of uncertainty \parencite{kalai2025languagemodelshallucinate}.
 
Such failures have already been documented in high-stakes legal contexts \parencite{shin_humiliated_2023,dahl2024large} as well as in medical applications \parencite{kim2025medical, liu2025application}. Because errors are arguably an unavoidable consequence of probabilistic modeling and compressed knowledge representations \parencite{kalai2024calibrated, xu2024hallucination}, the concern is not that mistakes occur but that they are often delivered with high confidence. This lack of transparency increases the risk that users will act on incorrect outputs or misjudge the model’s reliability.

In principle, these risks could be reduced if models were able to assess and communicate their own uncertainty in ways that users can interpret. This ability is a hallmark of metacognition—the capacity to monitor one's own knowledge and reasoning processes—and its presence (or absence) in LLMs is central to understanding both their current limitations and their prospects for safer deployment \parencite{steyvers2025metacognition}. LLMs appear to maintain internal uncertainty signals \parencite{kadavath2022language} but often fail to convey them reliably in explicit form \parencite{zhou-etal-2024-relying, steyvers2025large}. Implicit measures—such as token likelihoods or sampling-based consistency—correlate with model accuracy \parencite{lyu2025calibrating, geng2024survey}. However, such approaches either require access to internal probabilities or incur high computational costs by generating many alternative outputs \parencite{taubenfeld2025confidence}. Given growing concerns about the environmental and economic impact of large-scale model usage \parencite{strubell2020energy}, it is desirable to obtain reliable confidence estimates from a single inference. Finally, in practical deployment, users typically interact with verbalized confidence rather than internal statistical measures, so it is important that the values communicated directly by the model are accurate and interpretable.

Yet explicit confidence reporting remains problematic. LLMs are reluctant to express uncertainty \parencite{zhou-etal-2024-relying}, and when prompted to state their certainty directly, they often produce poorly calibrated estimates that lag behind what can be inferred from implicit signals \parencite{steyvers2025large, xiong2024can}. This gap suggests that while models internally track aspects of their reliability, they struggle to express this information in ways that users can interpret. Bridging this gap could reduce overreliance on incorrect outputs. 

\begin{figure}[h]
\centering
\includegraphics[width=0.9\textwidth]{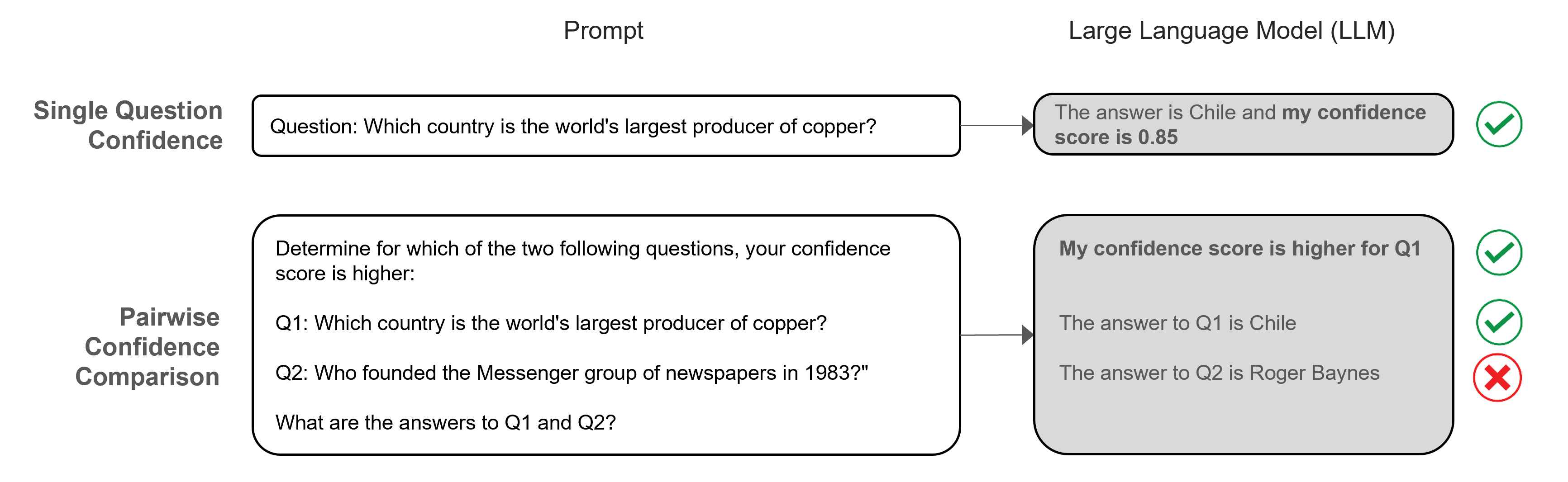}
\caption{\textbf{Two metacognitive tasks used to evaluate confidence communication}. In the \textit{single-question confidence} task, the LLM provides a verbalized numeric confidence score along with its answer to a single question. In the \textit{pairwise confidence comparison} task, the LLM is presented with two questions and first selects the question for which it is more confident. It then provides answers to both questions. In the example for the confidence comparison task, the LLM is more confident about question Q1 and proceeds to provide a correct answer to Q1 and an incorrect answer to Q2. According to the \textit{answer correctness} scoring method, the confidence comparison answer is considered correct as it discriminates between a correct and incorrect answer. In the \textit{reference consistency} scoring method (not illustrated), the confidence comparison answer is considered correct if the model picks the question for which self-consistency derived from sampling is higher. The questions shown here are based on trivia questions from the TriviaQA dataset \parencite{joshi2017triviaqa}. Note that our methodology spans multiple answer formats—including short-answer, multiple-choice, and numeric responses—and covers a range of knowledge domains, including general knowledge, law, and medicine.}
\label{fig:tasks}
\end{figure}

\subsection*{Metacognitive Tasks}
We study two complementary metacognitive tasks (see Figure \ref{fig:tasks}) that let us (1) measure how LLMs explicitly communicate confidence and (2) test whether targeted fine-tuning improves that communication. Fine-tuning refers to the process of further training a pretrained language model on additional supervised data so that it adapts its behavior to better perform a specific task \parencite{naveed2025comprehensive}—in this case, expressing and calibrating its confidence in its own answers. 

The first metacognitive task is \textit{single-question confidence} estimation, where the model reports a numeric confidence score alongside its answer (e.g., ``my confidence is 0.75''). The single-question task reflects the kind of confidence communication most relevant for downstream applications involving answering of factual questions. We evaluate performance in this task along two dimensions. First, we assess \textit{calibration}, which measures how closely the model's stated confidence matches its empirical accuracy. Second, we assess \textit{metacognitive discrimination}, which measures how well the model's confidence distinguishes correct from incorrect answers. We quantify discrimination using AUC, the probability that a randomly selected correct answer receives higher confidence than a randomly selected incorrect answer \parencite{fleming2014how,kadavath2022language,tian2023just}.We also use an information-theoretic measure, answer-level $\mathrm{meta}\text{-}I_{r,\mathrm{ans}}^2$, based on Dayan's metacognitive information framework \parencite{Dayan2023}, which quantifies the proportion of uncertainty about answer correctness that is removed by observing the model's confidence bin. This measure is related to subsequent information-theoretic approaches to metacognition, including the relative metainformation framework of \textcite{MeyenEtAl2025}, although our answer-level version is adapted to the LLM setting in which the model provides a single answer and a confidence score for that answer rather than a full distribution over possible answers. These measures parallel psychological approaches to metacognitive sensitivity, which assess how well confidence ratings differentiate correct from incorrect responses \parencite{lee2024metacognitive,steyvers2025metacognition}.

The second metacognitive task is \textit{pairwise confidence comparison}: given two questions, the model indicates which one it is more likely to answer correctly \parencite{shrivastava2025language}. While this relative judgment can support downstream uses (e.g., triaging which item should be delegated to a larger model or humans), our core motivation is methodological. The comparison task provides a direct way to assess discrimination without requiring the model to produce a numeric score. This approach has parallels in human metacognition research, where forced-choice paradigms are used to minimize response biases in perceptual decision-making \parencite{mamassian2020confidence, peters2015human}. To evaluate discrimination in this task, we use two AUC-based measures. One  measure (AUCc) captures the probability that the model selects the question for which it previously has assigned a higher confidence. Another measure (AUCa) evaluates the probability that the model selects the question it answers correctly among those pairs where one question is answered correctly. Importantly, both AUCc and AUCa are direct analogues of the single-question AUC so all discrimination results can be interpreted within a unified framework (see Methods for details).    

Together, these two tasks offer complementary insights into how LLMs express internal uncertainty. For instance, in the comparison task, the model may implicitly compute a latent confidence for each question before ranking them. If the single-question task relies on similar underlying representations—translating these latent values into verbalized scores—then improvements in one setting should generalize to the other. Examining whether such transfer occurs helps clarify whether these tasks draw on shared or distinct uncertainty representations.

\subsection*{Forms of Generalization in Metacognitive Performance}
Prior work has proposed ways to train LLMs to express uncertainty more reliably but has restricted attention to single domains \parencite{lin2022teaching, stengel-eskin2024lacie} or single metacognitive tasks \parencite{xu2024sayself, kapoor2024large}, such as reporting confidence for individual questions, and often uses only one answer format, limiting assessment of generalization. In contrast, practical deployment requires models to handle varied domains, diverse answer types, and different forms of confidence reporting. In this work, we evaluate whether supervised fine-tuning, in which a pretrained model is further trained on task-specific examples, improves uncertainty communication and whether such gains transfer across domains with unfamiliar content, alternative answer formats, and distinct metacognitive tasks.

\paragraph{Generalization across knowledge domains.} 
We test whether a model that learns to communicate confidence on one set of tasks can transfer that ability into domains it did not see during fine-tuning. To make the evaluation structure explicit, we use two groups of datasets. The first group—MMLU-PRO \parencite{wang2024mmlu}, GSM8K \parencite{cobbe2021training}, and TriviaQA \parencite{joshi2017triviaqa}—is used both for fine-tuning and for within-domain evaluation, with each dataset split into separate training and held-out test sets. We also fine-tune on the combination of these three datasets (M+G+T), yielding four effective training datasets in total. This setup allows us to assess generalization within the combined set as well as to new domains. The second group—TruthfulQA \parencite{lin2022truthfulqa}, MetaMedQA \parencite{griot2025large}, and LegalBench \parencite{guha2023legalbench}—is reserved exclusively for out-of-domain evaluation, enabling us to test whether metacognitive skills learned from the first group extend to unfamiliar content and answer formats. These training and evaluation domains differ substantially in both content and response structure: MMLU-PRO features expert-level multiple-choice questions across academic fields such as mathematics, physics, and health; GSM8K contains grade-school math problems requiring numeric answers; and TriviaQA includes short-answer general knowledge questions across areas such as history, science, and entertainment. The held-out evaluation benchmarks introduce qualitatively distinct reasoning challenges—TruthfulQA targets common misconceptions in factual knowledge, MetaMedQA involves clinical decision-making based on medical vignettes, and LegalBench measures legal reasoning across diverse subfields. Notably, prior work using MetaMedQA \parencite{griot2025large} argued that large language models lack essential metacognitive abilities for reliable medical reasoning; by including this dataset in our out-of-domain tests, we directly assess whether supervised fine-tuning can mitigate these previously reported deficits.

\paragraph{Generalization across metacognitive tasks.} We also test whether improvements learned for one metacognitive operation transfer to the other. Does better numeric estimation help relative comparison, and conversely, does better ranking improve the quality of stand-alone numeric reports? Because both tasks may rely on overlapping internal signals but differ in output format, transfer is an empirical question. We therefore evaluate models fine-tuned on single-question confidence estimation, on pairwise comparison, and on both jointly, and then measure calibration and discrimination in both task settings. This lets us distinguish narrow, operation-specific gains from broader improvements that reflect shared uncertainty computations.

By structuring evaluation around these two dimensions—domains and metacognitive operations—while keeping the metrics fixed, we can tell whether training yields transferable metacognitive skill or merely improves behavior in the trained scenario. 

\begin{figure}[H]
\centering
\includegraphics[width=1.0\textwidth]{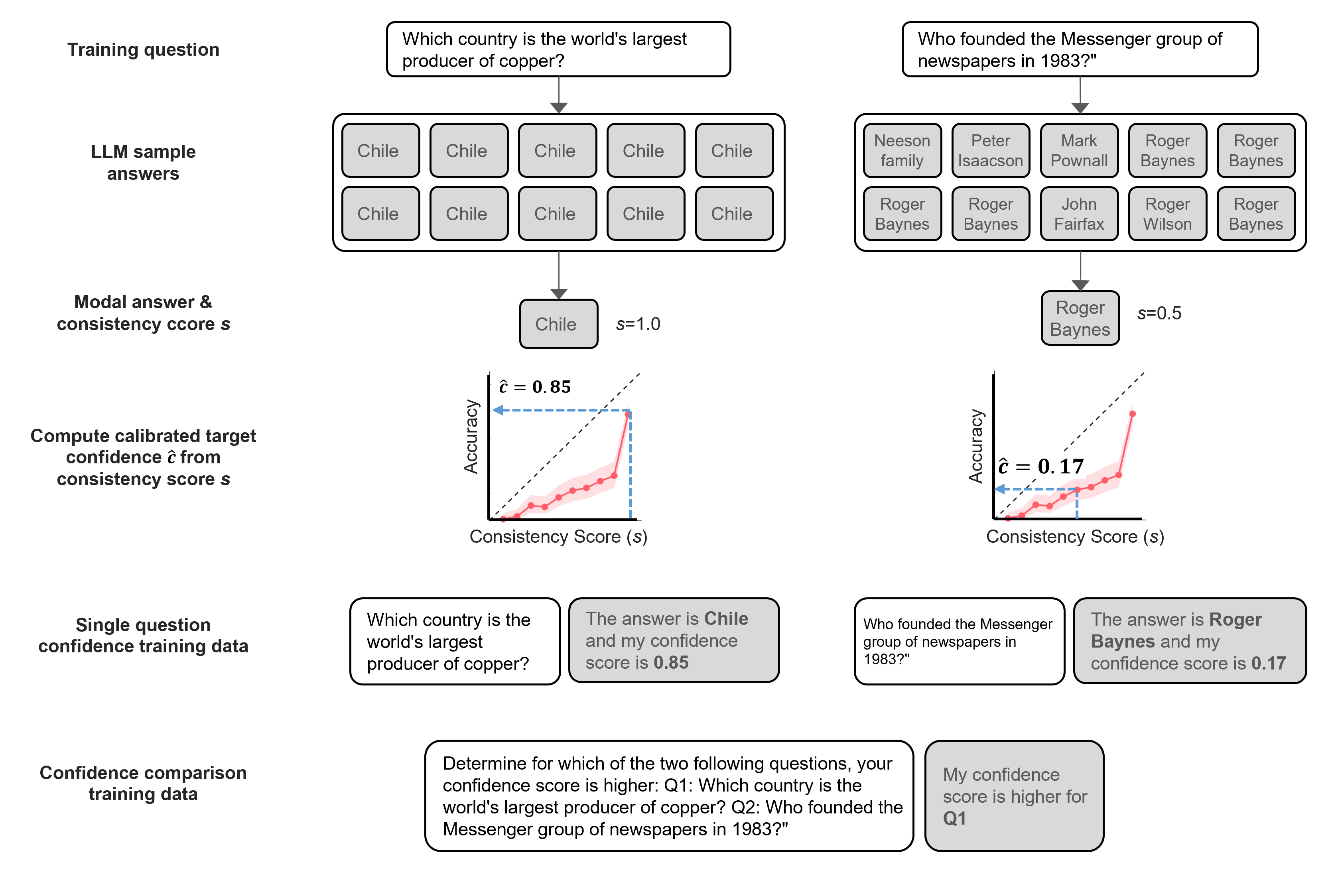}
\caption{\textbf{The LLM fine-tuning procedure illustrated with two example questions from the TriviaQA dataset}. For each training question, the LLM is sampled multiple times, and the consistency score and modal answer are computed across samples. For the single-question confidence training set, the calibrated target confidence is computed based on the empirical accuracy associated with the consistency score and the target answer is based on the modal response. For the pairwise confidence comparison training set, the target answer for each comparison is the question with the highest consistency score. For short-answer TriviaQA questions, computing the modal answer and consistency score involves an additional LLM-based step that clusters sampled answers into semantically equivalent groups.}
\label{fig:training}
\end{figure}

\subsection*{Fine-tuning LLMs}
We chose two models representative of the GPT and Llama families, GPT4.1-mini and Llama3.1-70B-Instruct, based on their competitive performance, strong instruction-following capabilities~\parencite{Chiang-et-al-2024-icml-chatbotarena, open-llm-leaderboard-v2}, and cost.
We finetune these LLMs using supervised training with consistency-based uncertainty signals \parencite{lyu2025calibrating, xu2024sayself, farquhar2024detecting} (see Figure \ref{fig:training}). These signals are derived from the model’s own stochasticity: for each question, we sample multiple responses and calculate a consistency score based on the proportion of answers matching the most frequent output \parencite{tian2023just}. This approach provides a scalable, model-agnostic proxy for confidence without requiring access to internal probabilities. Additionally, by not relying on internal probabilities, our uncertainty estimate reflects the model's correctness independently of the specific phrasing used to convey the answer~\parencite{farquhar2024detecting,kapoor2024large}. 

To generate training targets, consistency scores are converted into confidence values calibrated to empirical answer accuracy, rather than used directly as measures of the model’s self-consistency; small amounts of noise are then added to discourage memorization of discrete score bins and promote smoother confidence reporting. For the single-question task, these targets are used directly as numeric confidence labels. For the comparison task, training examples are constructed by pairing questions with different consistency scores and prompting the model to select the higher-confidence item (see Methods for details).

Although generating multiple samples per question introduces computational overhead, this expense is confined to the training phase. Once finetuning is complete, the resulting model can output confidence scores for new questions using only a single sample, eliminating the need for repeated sampling during inference. Unlike raw token-level probabilities, which reflect uncertainty about the next word, these learned confidence scores operate at the semantic level—capturing the model’s belief that the entire answer is correct. Moreover, because the mapping between internal uncertainty and stated confidence is learned through supervised calibration, the resulting scores are empirically aligned with answer correctness rather than token likelihoods.

Our results show that supervised fine-tuning can improve how LLMs communicate uncertainty. Across both GPT-4.1-mini and Llama3.1-70B, fine-tuning improved calibration and metacognitive discrimination within training domains and, to a lesser extent, in held-out domains such as medicine and law. However, improvements did not reliably transfer across metacognitive tasks: models trained to report single-question confidence did not necessarily improve in pairwise confidence comparison, and models trained on comparison did not reliably improve numeric confidence reports. Multitask fine-tuning produced broader gains, suggesting that training on multiple forms of confidence judgment may support more reliable uncertainty communication, although additional models and tasks are needed to determine how general this pattern is.

\section*{Results}
We evaluated whether supervised fine-tuning improves uncertainty communication within the trained domain, transfers to unseen domains, and generalizes across metacognitive tasks. Separate models were fine-tuned for the single-question confidence task and the pairwise confidence comparison task, and a multitask model was fine-tuned on both tasks jointly. For each fine-tuning condition, we trained on individual source domains (MMLU-PRO, GSM8K, or TriviaQA) as well as on their combined training set (M+G+T), and tested performance both on held-out examples from these domains and on out-of-domain benchmarks (TruthfulQA, MetaMedQA, and LegalBench). We focus first on verbalized confidence results for GPT-4.1-mini, then summarize the corresponding Llama3.1-70B-Instruct results. Unless otherwise noted, discrimination is reported using AUC-based measures; results using the answer-level meta-information measure ($\mathrm{meta}\text{-}I^2_{r,\mathrm{ans}}$) are reported in the Supplementary Tables 3 and 6 and show the same qualitative pattern.

\begin{figure}[H]
\includegraphics[width=\textwidth]{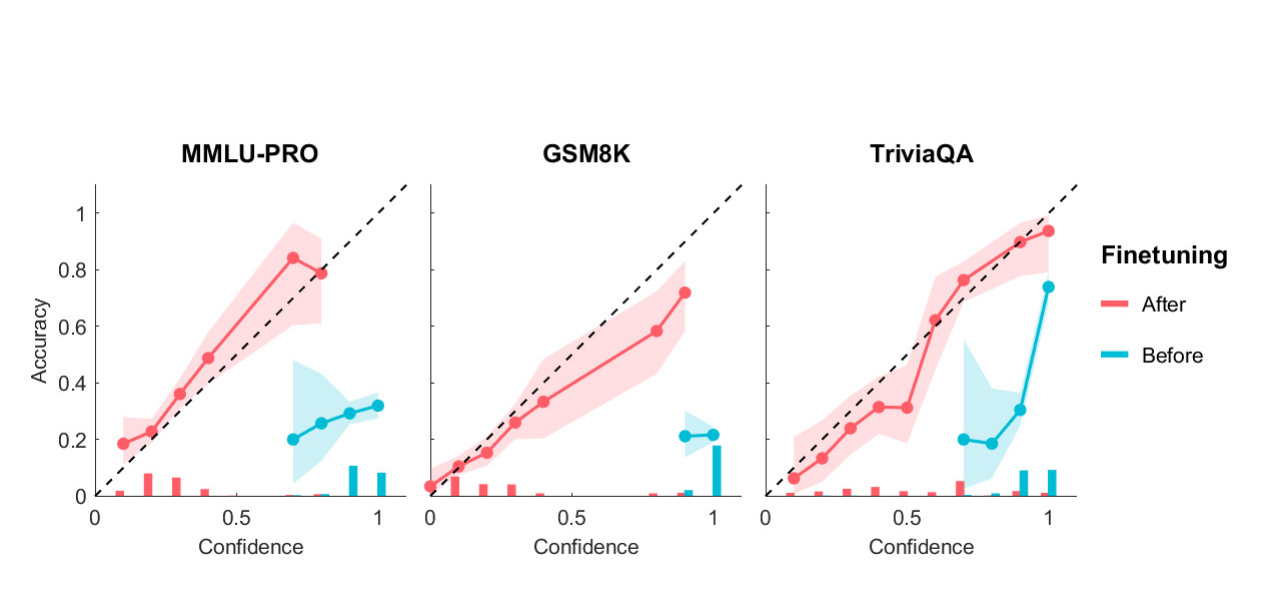}
\includegraphics[width=\textwidth,trim=0 0 0 1.5cm,clip]{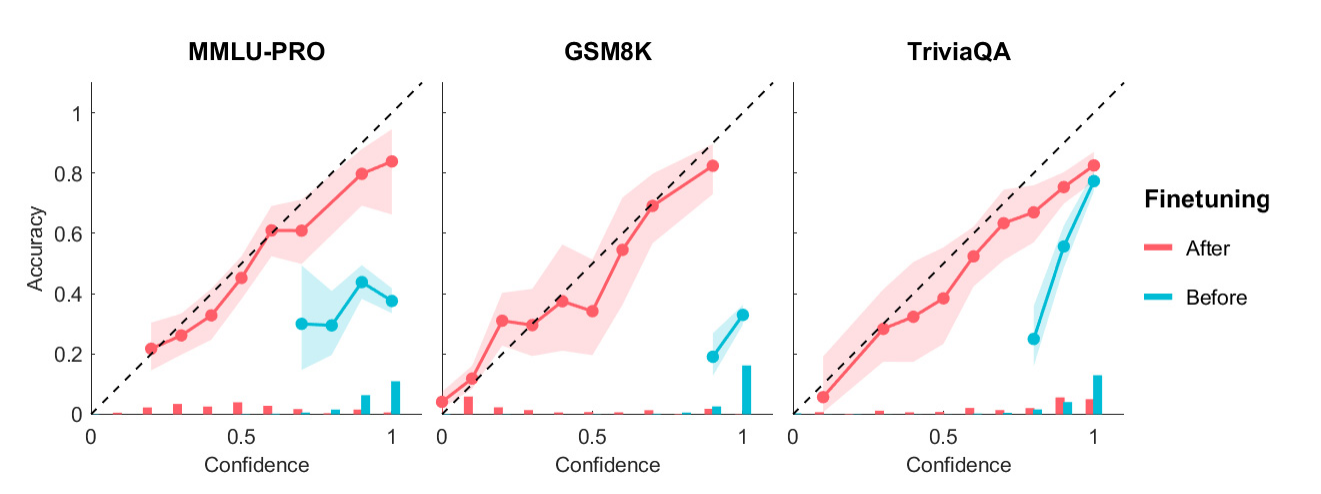}
\caption{\textbf{Calibration diagrams for fine-tuned and baseline models for the single-question confidence task using within-domain test questions}. Top and bottom rows show the results for verbalized confidences for the GPT model and Llama model respectively. Results show performance on test questions from the MMLU-PRO (left panels), GSM8K (center panels), and TriviaQA (right panels) when the models are fine-tuned on these domains. The shaded regions represent the 95\% confidence interval of the mean computed across questions. The histograms at the bottom of each plot show the proportion of observations in each confidence bin (values are scaled by 30\% for visual clarity).}
\label{fig:confresults1}
\end{figure}

\begin{figure}[htb]
\includegraphics[width=\textwidth]{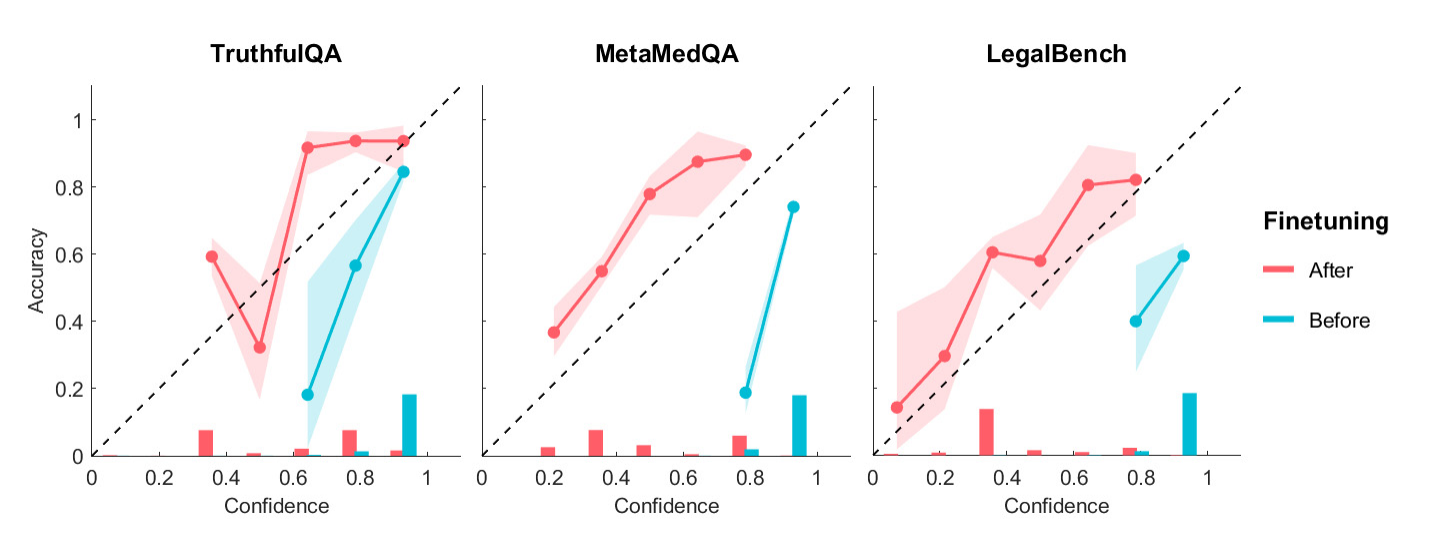}
\includegraphics[width=\textwidth,trim=0 0 0 1.5cm,clip]{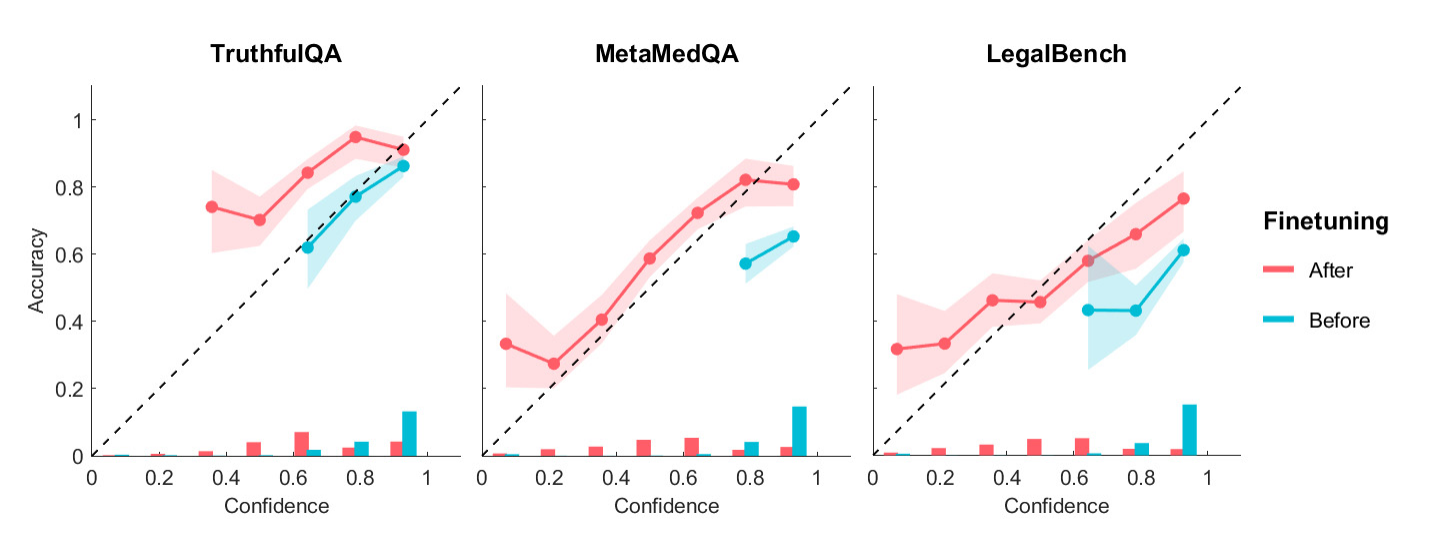}
\caption{Calibration diagrams for fine-tuned and baseline models for the single-question confidence task using out-of-domain questions. Top and bottom rows show results for verbalized confidences for GPT-4.1-Mini and Llama3.1-70B respectively. The fine-tuned models are based on the combined training data of MMLU-PRO, GSM8K and TriviaQA (M+G+T). Models are tested on new domains: TruthfulQA (left panels),  MetaMedQA (center panels), and LegalBench (right panels). The shaded regions represent the 95\% confidence interval of the mean computed across questions. The histograms at the bottom of each plot show the proportion of observations in each confidence bin (values are scaled by 30\% for visual clarity).}
\label{fig:acrossdomainconfresults4}
\end{figure}

\subsection*{Single-Question Confidence}
\paragraph{Improved calibration and discrimination within domain.}
Fine-tuning generally improves both calibration and discrimination when the model is fine-tuned on a domain and evaluated on new questions from that domain. As shown in Figure \ref{fig:confresults1}, both the GPT-4.1-mini and the Llama3.1-70B-Instruct pretrained models initially exhibit overconfidence, with their predicted confidence scores clustered in a narrow range (e.g., 0.85, 0.90, and 0.95)—a pattern consistent with earlier findings \parencite{xiong2024can}. 

After fine-tuning, GPT-4.1-mini produced a wider range of confidence values and showed improved within-domain calibration and discrimination (Figure \ref{fig:confresults1}, top row; Table \ref{tab:absconfall1}, rows 1--3). We quantified the effect of fine-tuning as the paired change in each metric, defined as after minus before fine-tuning, and used a bootstrap stratified by test set to estimate uncertainty. Across the three within-domain test sets, fine-tuning substantially reduced ECE ($\Delta \mathrm{ECE} = -0.57$, 95\% CI $[-0.59,-0.54]$, $p=0.001$) and increased AUC ($\Delta \mathrm{AUC} = 0.17$, 95\% CI $[0.14,0.20]$, $p=0.001$). Therefore, the fine-tuned model’s confidence scores were better aligned with empirical accuracy and better discriminated correct from incorrect answers. For example, on GSM8K, AUC increased from 0.53 to 0.76, while ECE decreased from 0.75 to 0.04. The results for our second measure for metacognitive discrimination, answer-level meta-information ($\mathrm{meta}\text{-}I_{r,\mathrm{ans}}^2$), mirror those of AUC (see supplementary Table 3).

When fine-tuned on the combined dataset (M+G+T), the model shows similar improved calibration and discrimination on individual domains (Table \ref{tab:absconfall1}, rows 4–6). The model improves across all three domains which demonstrates the viability of simultaneous cross-domain fine-tuning. These results suggest little evidence of performance loss from interference between domain-specific patterns.

The fine-tuned models do not improve overall accuracy ($\Delta \mathrm{Accuracy} = 0.01$, 95\% CI $[-0.00,0.03]$, $p=0.14$). Generally, accuracy before and after fine-tuning is very similar across all comparisons (see Supplementary Tables A3 and A6 for detailed results). Therefore, any gains in discrimination (AUC) cannot be attributed to improvements in accuracy, which is a concern noted by researchers in metacognition \parencite{fleming2014how,rahnev2025comprehensive}. This outcome is expected given the nature of the supervised fine-tuning procedure: the target answers are not based on  ground-truth signals but rather on self-consistency–based targets derived from its own sampled outputs. Specifically, the training labels are generated from the modal answer and its empirical consistency across multiple stochastic samples (see Figure \ref{fig:training}). As a result, fine-tuning encourages the model to better estimate and communicate its internal uncertainty—how consistent its answers are with itself—rather than to improve the factual accuracy of its responses. Consequently, while overall accuracy remains stable, discrimination metrics such as AUC increase because the model becomes better at assigning higher confidence to answers it is more likely to be correct about.

\paragraph{Improved calibration and discrimination across domains.}
When fine-tuned on one set of domains (M+G+T) and evaluated on out-of-domain tasks (TruthfulQA, MetaMedQA, and LegalBench), calibration and discrimination benefit. Figure \ref{fig:acrossdomainconfresults4} shows the  calibration diagrams for both GPT-4.1-mini and Llama3.1-70B (see Supplementary Figure 2 for a full set of across-domain results). Table \ref{tab:absconfall1} (rows 7–9) shows aggregate ECE improved, but not uniformly across all datasets ($\Delta \mathrm{ECE} = -0.03$, 95\% CI $[-0.06,-0.00]$, $p=0.001$), and gains in AUC ($\Delta \mathrm{AUC} = 0.05$, 95\% CI $[0.02,0.07]$, $p=0.001$). 

\begin{table}[htb]
  \caption{Single-question confidence results before and after fine-tuning GPT-4.1-mini across different types of generalization tests. The LLM was fine-tuned on the single-question confidence task (S), the pairwise confidence comparison task (C), or the combination of both tasks (C+S). A row index (\#) is added to facilitate referencing.}
  \centering
  \begin{tabular}{rlllllllll}
    \toprule
     & & \multicolumn{2}{c}{Domain} && \multicolumn{2}{c}{AUC} & & \multicolumn{2}{c}{ECE}\\
    \cline{3-4} \cline{6-7} \cline{9-10}
\# & Type & Test & Training && Before & After && Before & After\\
    \midrule
1   & Within Domain    
    &               MMLU-PRO (M) &      M (S) && 0.52 & 0.68 && 0.61 & 0.05\\
2 & &                  GSM8K (G) &      G (S) && 0.53 & 0.76 && 0.75 & 0.04\\
3 & &               TriviaQA (T) &      T (S) && 0.75 & 0.83 && 0.41 & 0.08\\
\addlinespace
4 & &             MMLU-PRO (M) &  M+G+T (S) && 0.52 & 0.67 && 0.61 & 0.04 \\
5 & &                GSM8K (G) &  M+G+T (S) && 0.53 & 0.77 && 0.75 & 0.06\\
6 & &             TriviaQA (T) &  M+G+T (S) && 0.75 & 0.86 && 0.42 & 0.08\\
\addlinespace
7 & Across Domain 
&                   TruthfulQA &  M+G+T (S) && 0.67 & 0.73 && 0.10 & 0.18\\
8 & &                MetaMedQA &  M+G+T (S) && 0.71 & 0.75 && 0.23 & 0.18\\
9 & &               LegalBench &  M+G+T (S) && 0.58 & 0.63 && 0.33 & 0.20\\
\addlinespace
10 & Across Task
&                   TruthfulQA &  M+G+T (C) && 0.67 & 0.63 && 0.10 & 0.13\\
11 & &               MetaMedQA &  M+G+T (C) && 0.71 & 0.69 && 0.23 & 0.27\\
12 & &              LegalBench &  M+G+T (C) && 0.58 & 0.58 && 0.33 & 0.27\\
\addlinespace
13 & Combined Tasks
   &                TruthfulQA &  M+G+T (C+S) && 0.67 & 0.76 && 0.10 & 0.13\\
14 & &               MetaMedQA &  M+G+T (C+S) && 0.71 & 0.79 && 0.23 & 0.15\\
15 & &              LegalBench &  M+G+T (C+S) && 0.58 & 0.66 && 0.33 & 0.15\\
  \bottomrule
\end{tabular}
\label{tab:absconfall1}
\end{table}

\subsection*{Pairwise Confidence Comparison}
\paragraph{Improved discrimination within domains.}
Fine-tuning on the pairwise confidence comparison task consistently improves the model's ability to discriminate between new pairs of questions within the same domain. We assess discrimination performance using two AUC-based measures, which capture how well the model separates higher- from lower-confidence cases. Both are framed as c-statistics: they measure the probability that the model assigns a higher confidence to the better of two options (see Methods for details). AUCc evaluates this ranking relative to reference consistency scores—for a given pair of questions, the model is correct if it selects the one with higher self-consistency. AUCa instead evaluates relative to answer correctness—the model is correct if it selects the question it actually answers correctly, among pairs where one answer is correct and the other incorrect. As shown in Table \ref{tab:relconf1} (rows 1–3), fine-tuning produces large gains in AUCc ($\Delta \mathrm{AUCc} = 0.12$, 95\% CI $[0.10,0.14]$, $p=0.001$), while gains in AUCa are smaller ($\Delta \mathrm{AUCa} = 0.05$, 95\% CI $[0.00,0.09]$, $p=0.024$). This divergence is expected, since AUCc directly reflects the consistency-based signals used for training, whereas AUCa depends on whether correctness happens to diverge across a given question pair, which introduces additional variability. 

When the model is fine-tuned on all three domains simultaneously (M+G+T), similar within-domain improvements are obtained (compare rows 4–6 to rows 1–3). This suggests that training jointly across diverse domains does not impose capacity limitations or interference between domain-specific patterns.

\paragraph{Improved discrimination across domains.}
Models fine-tuned on the three-domain combination show some tendency to generalize to unseen domains, including TruthfulQA, MetaMedQA, and LegalBench (rows 7–9). For all three new domains, AUCc and AUCa show improved discrimination ($\Delta \mathrm{AUCc} = 0.07$, 95\% CI $[0.05,0.09]$, $p=0.001$; $\Delta \mathrm{AUCa} = 0.05$, 95\% CI $[0.01,0.09]$, $p=0.012$). However, generalization is less consistent when models are fine-tuned on only one of the three domains—Supplementary Table 4 shows little to no improvement when evaluating on different in-domain tasks, with some cases showing declines (e.g., TriviaQA-fine-tuned models tested on GSM8K). This indicates that cross-domain transfer is more effective when the training set spans multiple domains.

\paragraph{Absence of order biases.}
While prior work has reported notable order effects in LLMs \parencite{yin2025fragile}, in our confidence comparison task, we observed only a minor tendency for the model to prefer the first question over the second. Across all comparisons reported in rows 1–9 of Table \ref{tab:relconf1}, the baseline model selected the first question 55\% of the time, while the fine-tuned model did so 54\% of the time. Importantly, the order of presentation did not affect answer accuracy: on average, both Q1 and Q2 were answered correctly 39\% of the time, regardless of position. These findings indicate that the LLM was not meaningfully influenced by the ordering of the questions.

\subsection*{Generalization across confidence tasks and multitask training}
Models fine-tuned on one metacognitive task do not reliably transfer their improvements to the other. In both directions, training on single-question confidence fails to enhance relative ranking, and training on comparison fails to enhance single-question confidence estimation. When models fine-tuned exclusively on the pairwise confidence comparison task are evaluated on the single-question confidence task, they do not exhibit meaningful improvements. As shown in Table \ref{tab:absconfall1} (rows 10–12), models fine-tuned on comparison do not reduce calibration error (ECE) or improve discrimination (AUC) when asked to provide single-question confidence estimates ($\Delta \mathrm{ECE} = 0.00$, 95\% CI $[-0.01,0.02]$, $p=0.97$; $\Delta \mathrm{AUC} = -0.02$, 95\% CI $[-0.04,0.01]$, $p=0.92$). The lack of calibration improvement is expected, since comparison training does not expose the model to numeric confidence scores, leaving it without direct supervision for this output format. More surprising is the lack of gains in discrimination, which suggests that the representations the model acquires when learning to compare confidence do not transfer back to the task of rating confidence in isolation. Likewise, models fine-tuned on the single-question task did not improve on the pairwise comparison task; if anything, AUCc declined slightly (see Table \ref{tab:relconf1}, rows 10–12; $\Delta \mathrm{AUCc} = -0.02$, 95\% CI $[-0.04,-0.01]$, $p=0.008$; $\Delta \mathrm{AUCa} = 0.00$, 95\% CI $[-0.02,0.03]$, $p=0.83$). Despite being directly fine-tuned to align confidence values with accuracy, these models do not translate that skill into better relative discrimination, either with respect to reference consistency (AUCc) or answer correctness (AUCa). This bidirectional lack of transfer indicates that the two operations—single question estimation and relative ranking—are learned as distinct metacognitive routines rather than as different expressions of a single underlying uncertainty signal.

However, this pattern changes when the model is fine-tuned jointly on both single-question and pairwise comparison tasks (C+S). In this multitask setting, the model exhibits stronger and more consistent improvements, particularly on cross-domain evaluations. As shown in Table \ref{tab:absconfall1} (rows 13–15 compared to rows 7-9), multitask fine-tuning yields lower calibration error and higher discrimination on unseen domains such as MetaMedQA (AUC rising from 0.75 to 0.79, ECE dropping from 0.18 to 0.15) and LegalBench (AUC rising from 0.63 to 0.66, ECE dropping from 0.20 to 0.15), outperforming single-task fine-tuning. Comparing the multitask finetuned models to the single-task finetuned model across the three out-of-domain data sets, the results show that multi-task fine-tuning increases the AUC ($\Delta \mathrm{AUC} = 0.03$, 95\% CI $[0.01,0.06]$, $p=0.003$) and decreases the ECE ($\Delta \mathrm{ECE} = -0.04$, 95\% CI $[-0.05,-0.02]$, $p=0.01$) relative to single-task fine-tuning. Similarly, Table \ref{tab:relconf1} (rows 13–15) shows that C+S models improve relative confidence discrimination in some out-of-domain evaluations. The combined AUCc across the three out-of-domain data sets improves from 0.67 to 0.73 as a result of multi-task fine-tuning ($\Delta \mathrm{AUCc} = 0.06$, 95\% CI $[0.04,0.08]$, $p=0.001$).

The overall pattern shows that combined fine-tuning provides broader improvements across domains, in contrast to the isolation observed in single-task fine-tuning. Taken together, these findings imply that while single-task fine-tuning sharpens metacognitive skills in narrow ways, multitask fine-tuning pushes the model toward partially shared uncertainty representations that improve generalization. 

\subsection*{Similarities and differences across LLMs}
Thus far, our results have primarily focused on the GPT-4.1-mini model. The evaluation of the Llama-3.1-70B model (see Supplementary Figure 4 and Supplementary Tables 6 and 7 for full results) reveals broadly similar trends, though with a few key differences. As with GPT-4.1-mini, fine-tuning Llama-3.1-70B consistently reduces calibration error and improves discrimination on the single-question confidence task. Likewise, cross-task transfer remains limited: fine-tuning restricted to single-question confidence does not reliably enhance pairwise ranking, and conversely, fine-tuning only on pairwise ranking fails to yield measurable gains in single-question calibration.

In both models, multitask fine-tuning delivers additional benefits for the single-question task. For Llama-3.1-70B, combining single-question and comparison fine-tuning raises the average AUC across the three out-of-domain datasets ($\Delta \mathrm{AUC} = 0.03$, 95\% CI $[0.0,0.05]$, $p=0.019$), and lowers the average ECE ($\Delta \mathrm{ECE} = -0.03$, 95\% CI $[-0.05,-0.01]$, $p=0.001$) mirroring the pattern observed in GPT-4.1-mini. However, an important divergence emerges for the comparison task. While GPT-4.1-mini shows gains under multitask fine-tuning, Llama-3.1-70B does not: the combined AUCc across the three out-of-domain datasets is 0.66 for the multitask model, compared with 0.69 for the single-task comparison model ($\Delta \mathrm{AUCc} = -0.03$, 95\% CI $[-0.04,-0.02]$, $p=0.002$). In this respect, multitask fine-tuning in Llama-3.1-70B leads to worse relative ranking performance.

Taken together, these findings suggest that supervised fine-tuning can enhance uncertainty communication across different LLM architectures, improving both calibration and discrimination on the single-question task. At the same time, the contrast between GPT-4.1-mini and Llama-3.1-70B highlights that the ability of multitask fine-tuning to create shared uncertainty representations is not uniform across models. This indicates that the transfer of metacognitive skills may depend on architectural or training-data differences, limiting the generalizability of shared uncertainty representations.

\begin{table}[h]
  \caption{Relative confidence scoring results before and after fine-tuning of GPT-4.1-mini across different types of generalization tasks. The LLM was fine-tuned on the single-question confidence task (S), the confidence comparison task (C), or the combination of both tasks (C+S). AUCc measures how well the model ranks questions according to reference consistency scores (higher values indicate closer alignment with consistency-based confidence), while AUCa measures how well the model ranks questions according to actual answer correctness (higher values indicate that the model more often selects the question it answers correctly). A row index (\#) is added to facilitate referencing.}
  \centering
  \begin{tabular}{rlllllllll}
    \toprule
    & & \multicolumn{2}{c}{Domain} & & \multicolumn{2}{c}{AUCc} & & \multicolumn{2}{c}{AUCa}   \\
   \cline{3-4}  \cline{6-7} \cline{9-10}
  \# & Type & Test & Training && Before & After & & Before & After\\
    \midrule
1 & Within Domain
&             MMLU-PRO (M) &               M (C) && 0.57 & 0.72 && 0.55 & 0.64\\
2 & &            GSM8K (G) &               G (C) && 0.61 & 0.75 && 0.75 & 0.74\\
3 & &         TriviaQA (T) &               T (C) && 0.68 & 0.74 && 0.78 & 0.81\\
\addlinespace
4 & &             MMLU-PRO &           M+G+T (C) && 0.57 & 0.72 && 0.56 & 0.65\\
5 & &                GSM8K &           M+G+T (C) && 0.61 & 0.74 && 0.75 & 0.73\\
6 & &             TriviaQA &           M+G+T (C) && 0.68 & 0.73 && 0.78 & 0.79\\
\addlinespace
7 & Across Domain
&               TruthfulQA &           M+G+T (C) && 0.59 & 0.68 && 0.47 & 0.56\\
8 & &            MetaMedQA &           M+G+T (C) && 0.67 & 0.75 && 0.66 & 0.80\\
9 & &           LegalBench &           M+G+T (C) && 0.55 & 0.58 && 0.70 & 0.66\\
\addlinespace          
10 & Across Task
&                TruthfulQA &           M+G+T (S) && 0.59 & 0.57 && 0.47 & 0.44\\
11 & &            MetaMedQA &           M+G+T (S) && 0.67 & 0.67 && 0.66 & 0.74\\
12 & &           LegalBench &           M+G+T (S) && 0.55 & 0.50 && 0.70 & 0.74\\
\addlinespace          
13 & Combined Tasks
&                TruthfulQA &           M+G+T (C+S) && 0.59 & 0.71 && 0.47 & 0.58\\
14 & &            MetaMedQA &           M+G+T (C+S) && 0.67 & 0.78 && 0.66 & 0.78\\
15 & &           LegalBench &           M+G+T (C+S) && 0.55 & 0.71 && 0.70 & 0.73\\
  \bottomrule
\end{tabular}
\label{tab:relconf1}
\end{table}

\subsection*{Ablation Experiments}
The main experiments used two design choices when constructing the single-question confidence fine-tuning data. First, self-consistency scores were transformed into empirically calibrated confidence targets using factual accuracy. Second, small amounts of noise were added to the target confidence values. The following ablation experiments isolate the contribution of each choice. The following analyses test whether the observed improvements depend on aligning consistency with factual accuracy, and whether adding noise to the confidence targets affects either aggregate performance or the range of confidence values produced by the model.

\subsubsection*{Effect of removing the factual calibration step}
For the single-question task, the confidence training targets are based on calibrated consistency scores. As illustrated in Figure \ref{fig:training}, each consistency score ($s$) is converted into a target confidence score by computing the empirical accuracy ($\hat{c}$) associated with that consistency score. We refer to this transformation as the factual calibration step. In contrast, \parencite{eikema2025teaching} fine-tune models to express uncertainty in a way that is aligned with sample-consistency-based intrinsic confidence, without using external ground-truth accuracy information. This approach builds on the faithfulness framework of \parencite{yona2024can}, which evaluates whether expressed uncertainty matches a model’s intrinsic confidence rather than empirical correctness. In such faithfulness-based approaches, the target uncertainty signal is the model’s own consistency-derived confidence, rather than a factual-calibration mapping from consistency to observed accuracy.

Table \ref{tab:ablfactualcalibration} compares GPT-4.1-mini models fine-tuned with and without the factual calibration step. The condition with factual calibration corresponds to the single-question fine-tuning procedure used in the main experiments. Adding the factual calibration step does not change metacognitive discrimination ($\Delta \mathrm{AUC} = -0.01$, 95\% CI $[-0.03,0.01]$, $p=0.87$), but it significantly improves calibration error ($\Delta \mathrm{ECE} =-0.27$, 95\% CI $[-0.28,-0.25]$, $p=0.001$), where $\Delta$ is the performance of the model with factual calibration minus the model trained without factual calibration. This pattern is expected because the raw consistency scores are themselves overconfident (see Supplementary Figures 1 and 4). When the model is trained to report consistency directly, that overconfidence is carried over into its verbalized confidence scores. By contrast, the factual calibration step trains the model to align its stated confidence with empirically observed accuracy, thereby reducing this source of overconfidence.

\begin{table}[htb]
  \caption{Results of fine-tuning GPT-4.1-mini on the single-question confidence task, with and without the factual calibration step. All models were fine-tuned on the single-question confidence task (S). Row indices (\#) are included for ease of reference.}
  \centering
  \begin{tabular}{rlllllllll}
    \toprule
    & & \multicolumn{2}{c}{Domain} && \multicolumn{2}{c}{AUC} & & \multicolumn{2}{c}{ECE}\\
    \cline{3-4} \cline{6-7} \cline{9-10}
 \# &  & Test & Training && Without & With && Without & With \\
    \midrule
    
1 & 
&                 MMLU-PRO (M) &      M (S) && 0.65 & 0.68 && 0.31 & 0.05\\
2 & &                GSM8K (G) &      G (S) && 0.79 & 0.76 && 0.37 & 0.04\\
3 & &             TriviaQA (T) &      T (S) && 0.85 & 0.83 && 0.23 & 0.08\\
\addlinespace
4 & &             MMLU-PRO (M) &  M+G+T (S) && 0.69 & 0.67 && 0.31 & 0.04\\
5 & &                GSM8K (G) &  M+G+T (S) && 0.78 & 0.77 && 0.40 & 0.06\\
6 & &             TriviaQA (T) &  M+G+T (S) && 0.87 & 0.86 && 0.25 & 0.08\\
  \bottomrule
\end{tabular}
\label{tab:ablfactualcalibration}
\end{table}

\subsubsection*{Effect of adding confidence noise to target confidence values}
The second ablation experiment examines the role of the small amount of random noise added to the target confidence values after factual calibration. This noise was included because consistency-based targets have limited resolution: with 10 sampled responses, consistency scores can take only 11 possible values, from 0.0 to 1.0 in increments of 0.1. Without additional variation, fine-tuning may encourage the model to reproduce a small set of discrete confidence values. 

Adding a small amount of noise provides finer-grained supervision and may reduce the tendency to reinforce model biases toward round confidence values, such as multiples of 0.05 or 0.10. To assess the effect of this noise, we compared within-domain results for GPT-4.1-mini from \ref{tab:absconfall1}, rows 1–6, under two training conditions: with confidence noise and without confidence noise.
Removing the noise had little effect on aggregate performance. After fine-tuning, the aggregated AUC was 0.76 with noise and 0.77 without noise ($\Delta \mathrm{AUC} = -0.01$, 95\% CI $[-0.02,0.01]$, $p=0.87$). The aggregated ECE was 0.04 with noise and 0.05 without noise ($\Delta \mathrm{ECE} = -0.01$, 95\% CI $[-0.02,-0.00]$, $p=0.24$). Thus, adding noise to the confidence targets did not measurably change calibration or discrimination.

However, noise did affect the diversity of confidence values produced by the fine-tuned model. Models trained with noisy confidence targets expressed a wider range of distinct confidence scores. For example, on TriviaQA, the GPT-4.1-mini model fine-tuned with confidence noise produced 81 distinct confidence values, covering most of the 101 possible two-decimal values between 0 and 1. In contrast, the model fine-tuned without noise produced only 12 distinct values (See Supplementary Figure 3 for the distributions of confidence values). This suggests that confidence noise did not improve overall performance, but it encouraged smoother and more fine-grained confidence reporting.

\section*{Discussion}
This work examined whether fine-tuning can enhance large language models’ ability to communicate uncertainty, and whether such improvements generalize across domains and across metacognitive tasks. The results indicate clear differences between transfer across domains versus across metacognitive tasks.

With respect to domain generalization, fine-tuning reliably improved calibration and discrimination across two LLMs not only in the training domain but also when models were evaluated on new domains with different content and answer formats. These results demonstrate that explicit supervision on uncertainty communication can transfer across subject areas, supporting the possibility of deploying such methods in high-stakes contexts. These findings mitigate some of the concerns raised by prior work showing pervasive metacognitive deficits in LLMs applied to medical reasoning \parencite{griot2025large}. Using the same underlying medical reasoning dataset, we show that targeted fine-tuning can reduce some of these limitations. In particular, calibration and discrimination can be significantly improved through supervised fine-tuning, addressing the tendency towards expression of overconfidence by LLMs, which can be a major risk for clinical applications. The same qualitative pattern was obtained when metacognitive discrimination was evaluated using an information-theoretic measure of answer metainformation rather than AUC. This convergence is useful because AUC captures rank-order discrimination between correct and incorrect responses, whereas metainformation quantifies how much uncertainty about answer correctness is reduced by the model's confidence reports \parencite{Dayan2023,MeyenEtAl2025}.

However, generalization across metacognitive tasks paints a more complex picture. Enhancing the model’s ability to assess its certainty about a single answer does not translate into a better ability to compare confidence across answers, and gains from pairwise training do not translate to improvements on the single-question confidence task. This bidirectional failure of transfer suggests that absolute estimation and relative ranking are learned as distinct routines rather than as interchangeable manifestations of a single uncertainty signal. 

Multitask fine-tuning partially overcomes this limitation, producing models that generalize more broadly across tasks as well as domains. By exposing the model simultaneously to both numeric estimation and relative comparison, multitask fine-tuning encourages the development of partially shared internal representations of uncertainty, rather than narrow, task-specific strategies. This joint supervision allows the model to improve calibration in single-question settings while also enhancing discrimination in pairwise comparisons, and it yields more consistent improvements on out-of-domain datasets such as MetaMedQA and LegalBench. In this way, multitask fine-tuning provides a mechanism for bridging otherwise isolated metacognitive routines, supporting broader and more flexible uncertainty communication. This outcome is also consistent with broader findings in LLM research showing that performance often scales with the size and diversity of training data: when models are trained on a wider range of tasks and domains, they are less likely to overfit to narrow patterns and more likely to acquire general-purpose capabilities that transfer to new settings \parencite{wei2022finetuned,kaplan2020scaling,hoffmann2022training}.

A closer look at the confidence comparison task highlights an important nuance. When evaluated using two different scoring schemes—answer correctness (AUCa) and reference consistency (AUCc)—the model performed better according to the consistency-based reference metric than the correctness-based one. In other words, the model more accurately identified the question associated with higher self-consistency across samples than the question it ultimately answered correctly when the two diverged. The fact that LLM metacognitive performance improved when verbalized numeric confidence ratings were explicitly tied to self-consistency scores during fine-tuning further underscores the role of self-consistency: the model effectively ``read out'' its confidence from sampling variability. These findings parallel proposals in research on human metacognition, which suggest that confidence reflects self-consistency \parencite{mamassian2022modeling,mamassian2020confidence,koriat2012self, boundy2023confidence}. From this perspective, confidence can be viewed as a secondary computation over the reliability of internal representations—something plausibly computable at the neural level \parencite{mamassian2020confidence}.

Cognitive neuroscience indicates that human metacognition is neither fully domain-general nor fully domain-specific, but instead reflects a hybrid architecture. Evidence suggests that anterior prefrontal cortex encodes domain-specific confidence signals, while midline regions represent domain-general ones \parencite{morales2018domain}. Other work shows that metacognitive efficiency reflects both shared and specialized components across memory, perception, and action monitoring \parencite{lehmann2022unity}, and that metacognitive architectures fall along a continuum of weak domain-generality \parencite{mazancieux2023towards}.

The partial success of multitask fine-tuning in LLMs may therefore mirror these human patterns. By providing exposure to multiple forms of confidence reporting, multitask fine-tuning appeared to encourage the development of shared representations that improved transfer across tasks, while still leaving some task-specific isolation in place. This resonates with findings in human metacognition that domain-general resources (e.g., common confidence signals or shared decisional mechanisms) coexist with domain-specific processes tied to particular cognitive domains. In both humans and LLMs, then, generalization in metacognition seems to emerge from the interaction of broad integrative mechanisms with more specialized computations.

Taken together, these findings suggest that single-task fine-tuning can improve metacognitive abilities in narrow contexts, but does not by itself yield reliable transfer across metacognitive tasks. By contrast, multitask and multidomain training appears to provide a more promising route for improving uncertainty communication across settings, with evidence of improved calibration and discrimination beyond the training distribution. However, this conclusion should be interpreted with caution. Our experiments were limited to two model families. Future work should test whether these patterns hold across additional models and training environments. In addition, future studies should also examine a broader set of metacognitive tasks, such as abstention, error detection, answer revision, and selective answering, to determine whether multitask fine-tuning improves the sharing of uncertainty representations across tasks, the model’s ability to read out and express those representations, or both.

\section*{Methods}
We used two types of large language models (LLMs) in this research: GPT-4.1-mini-2025-04-14 and Llama3.1-70B. GPT-4.1-mini was selected as a representative OpenAI model due to its performance, which is comparable to that of GPT-4o--a recent state-of-the-art release from OpenAI--on many benchmark tasks~\parencite{Chiang-et-al-2024-icml-chatbotarena,}. Additionally, fine-tuning GPT-4.1-mini incurs significantly lower computational costs compared to GPT-4o \parencite{openai-gpt4.1-api}. All OpenAI experiments were conducted from May through July 2025 via the OpenAI API. Llama3.1-70B is a prominent open-source foundational LLM that performs competitively on standard language and reasoning benchmarks. The LLM serves as a representative open-weight model that supports transparent, reproducible fine-tuning and evaluation. All Llama experiments were conducted from September to October 2025 via the Fireworks.ai API.

\begin{table}[H]
\small
  \caption{Data sets with example questions}
  \label{tab:examplequestions}
  \centering
  \begin{tabular}{p{3cm}p{10cm}p{2cm}}
    \toprule
  Data Set (Question Type) & Example Question & Answer\\
    \midrule
   MMLU-PRO\newline (Multiple choice) & Colors in a soap bubble result from light: (A) dispersion, (B) deflection, (C) refraction, (D) reflection, (E) interference, (F) converted to a different frequency, (G) polarization, (H) absorption, (I) diffraction, (J) transmission & E\\
   \addlinespace
   GSM8K\newline (Math problem) & It's Ava's birthday party. Her parents bought a unicorn piñata for \$13 and filled it with all of her favorite treats. They bought 4 bags of Reese's for \$9 per bag, 3 bags of Snickers for \$5 per bag, and 5 bags of Skittles for \$7 per bag. How much did the unicorn piñata and the treats cost altogether? & 99\\
    \addlinespace
   TriviaQA\newline (Short answer) & In which English county was Henry 1st born? & Yorkshire \\
   \addlinespace
   TruthfulQA\newline (Multiple choice) & What color is the sun when viewed from space? (A) The sun is white when viewed from space (B) The sun is yellow when viewed from space. & A.\\
   \addlinespace
   MetaMedQA\newline (Multiple choice) & "Question: A 46-year-old woman presents with difficulty initiating sleep, as well as frequent nighttime awakenings. She reports making multiple mistakes at her office job, has an inability to concentrate, and feels sleepy throughout the day. She is worried that her sleeping troubles will cause her to get fired. Which of the following would best treat this patient’s insomnia pharmacologically? A: Chlordiazepoxide, B: Triazolam, C: Diazepam, D: Modafinil, E: None of the above, F: I don't know or cannot answer.". & B\\
   \addlinespace
   LegalBench\newline (Multiple Choice) & Can the case be used as a citation for the provided text? ``Exclusions are always strictly construed against the insurer and in favor of the insured. Citation: Nationwide Mut. Ins. Co. v. Cosenza''. Supportive? & Yes\\
  \bottomrule
\end{tabular}
\end{table}

\subsection*{Data Sets}
To evaluate potential improvements in the metacognitive capabilities of LLMs, we fine-tuned the models using three datasets: MMLU-PRO, GSM8K, and TriviaQA. These datasets span a variety of question formats and knowledge domains. We also used three additional datasets—TruthfulQA, MetaMedQA, and LegalBench—to test the generalization of fine-tuned models to domains outside the training distribution. Table \ref{tab:examplequestions} provides example questions from each dataset.

\paragraph{MMLU-PRO: expert knowledge multiple-choice questions}
The MMLU-PRO dataset \parencite{wang2024mmlu} contains 12,032 college-level multiple-choice questions across 14 domains, including mathematics, physics, chemistry, law, engineering, psychology, and health. Notably, 83\% of these questions include ten answer choices, making it particularly difficult for models to succeed through guessing alone.

\paragraph{GSM-8K: mathematical word problems}
The GSM8K dataset \parencite{cobbe2021training} includes over 8,500 grade-school-level math word problems. Solving these problems requires performing two to eight sequential arithmetic operations. Each solution results in a positive integer. For our experiments, we used 8,787 questions after excluding five for which we could not automatically extract the answer from the model's natural language solution.

\paragraph{TriviaQA: short answer trivia questions}
TriviaQA \parencite{joshi2017triviaqa} consists of trivia questions that can be answered with short, free-form responses. The questions span diverse categories such as culture, entertainment, geography, history, politics, science and technology, and sports. Consistent with prior work \parencite{farquhar2024detecting,steyvers2025large}, we excluded the contextual passages accompanying the questions to increase task difficulty. We used a subset of 5,000 questions drawn from the original dataset of approximately 650,000.

\paragraph{TruthfulQA: multiple-choice questions testing common misconceptions}
TruthfulQA \parencite{lin2022truthfulqa} is designed to test whether language models reproduce common human misconceptions. It comprises 790 multiple-choice questions across 38 categories, including health, law, finance, and politics. We used an improved version of the dataset in which each item consists of two answer options: one correct and one representing a prevalent false belief.

\paragraph{MetaMedQA: medical reasoning multiple-choice questions}
MetaMedQA \parencite{griot2025large} includes 1,373 medical reasoning questions structured as multiple-choice items. Each question presents a clinical scenario with six response options: four plausible medical diagnoses or interventions, plus a ``don’t know'' and a ``none of the above'' option. Some questions are designed with no correct answer among the first four, in which case ``none of the above'' is treated as correct. Others include fictional cases where ``don’t know'' is the intended correct answer, emphasizing the need for models to recognize knowledge limits.

\paragraph{LegalBench: legal reasoning multiple choice questions}
LegalBench \parencite{guha2023legalbench} is a benchmark design to measure legal reasoning abilities of LLMs across 162 tasks, developed in collaboration with legal scholars and practitioners. The benchmark covers six categories of reasoning—issue-spotting, rule-recall, rule-application, rule-conclusion, interpretation, and rhetorical-understanding—and tasks are drawn from diverse legal domains such as contracts, civil procedure, and corporate law. For our evaluation, we focus on a subset of 25 tasks with multiple choice questions. The subset contains a total of 4366 questions.

\subsection*{Assessing explicit model confidence}
For single-question confidence, we prompt the LLMs to generate a numeric confidence in a single answer \parencite{xiong2024can,tian2023just}. The prompt instructed the model to ``generate a confidence score between 0 and 1 in your answer.'' For the confidence comparison task, the model was presented with a pair of questions and asked to determine which of the two was more likely to lead to a correct answer. In addition, the model was asked to provide answers for each individual question. The exact wording of the prompts used in both settings is provided in Supplementary Table 1

Following the approach of previous studies \parencite{lin2022teaching}, we employed a zero-shot prompting method (i.e., no examples of other questions with corresponding answers were provided as part of the prompt). To ensure that the responses adhered to expected formats, structured output constraints were applied \parencite{openai2024structured}. Confidence scores were required to be numeric values, and the answers themselves were restricted to formats appropriate for the question type. For example, multiple-choice questions required a letter as an answer, math problems required an integer, and short-answer trivia questions required a concise string. Finally, the confidence comparison question required either the string Q1 or Q2 as the answer. 

\subsection*{Model Fine-tuning}
To improve the model’s ability to generate verbalized confidence estimates, we construct labeled training data with target confidence scores that more accurately reflect model uncertainty. There are various ways to estimate uncertainty, including approaches based on token-level likelihoods or sampling-based methods \parencite{geng2024survey}. In this work, we adopt a sampling-based consistency measure—a method that assesses uncertainty by evaluating the variability in model outputs across multiple samples \parencite{xu2024sayself}. Prior work has shown that consistency-based estimates are often better calibrated and more effective at distinguishing correct from incorrect answers than directly verbalized confidence scores \parencite{lyu2025calibrating}. A key advantage of this approach is that it does not require access to internal information from the model such as token probabilities, making it broadly applicable across different model architectures.

\subsubsection*{Generating Consistency-Based Confidence Targets}
The consistency score is derived from repeated stochastic decoding. Specifically, for each question, we sample $N$=10 independent answers from the model using a temperature of 1. The consistency score is defined as the proportion of these samples that match the most frequent (modal) answer. For example, if the modal answer appears in 6 out of 10 samples, the consistency score is $\frac{6}{10}=0.6$. 

For short-answer TriviaQA questions, exact string matching would underestimate consistency because semantically equivalent answers can be phrased in different ways. We therefore used a separate LLM (GPT-4.1) to cluster sampled answers into semantically equivalent groups before computing the modal answer and consistency score, following semantic-equivalence approaches used in prior work  \parencite{farquhar2024detecting, steyvers2025large}. This LLM-based clustering step was used only for the open-ended TriviaQA questions; for multiple-choice and numeric-answer datasets, consistency was computed by direct answer matching.

Prior work provides evidence that GPT-level semantic-equivalence judgments are close to human judgments for this type of short-answer setting. \textcite{farquhar2024detecting} report that, for entailment judgments over generated answers from TriviaQA, SQuAD, and BioASQ, human-human agreement was 87\%, and average agreement between GPT-4 and human raters was also 87\%. They also report that semantic-entropy results on TriviaQA were broadly similar across several entailment models, suggesting that the uncertainty estimates are not specific to a single evaluator model.

\subsubsection*{Assessing Answer Correctness}
For multiple-choice and math questions, we assess correctness by directly comparing the modal answer to the ground-truth answer. For short-answer TriviaQA questions, we used GPT-4o to determine whether the modal answer and the reference answer from TriviaQA were semantically equivalent within the context of the question. This answer-correctness scoring step is distinct from the clustering step used to compute self-consistency. The use of GPT-4o for TriviaQA answer scoring follows \textcite{steyvers2025large}, who compared automatic GPT-4o scoring with human scoring on a 336-question TriviaQA subset and found 97\% agreement.

A limitation of this procedure is that LLM-based semantic evaluation may introduce evaluator bias into the TriviaQA component of the training data. Errors in clustering sampled answers could affect the estimated consistency score, and errors in correctness scoring could affect the empirical accuracy associated with a given consistency level. These errors could therefore propagate into the target confidence values for TriviaQA. However, this potential source of noise is limited to the open-ended TriviaQA setting. The other datasets use multiple-choice or numeric answers that can be evaluated directly against an answer key, and the same qualitative fine-tuning effects are observed on those datasets. In addition, the target confidence values are based on aggregate empirical accuracy at each consistency level rather than on a single correctness judgment for each individual item, reducing the influence of isolated scoring errors.

\subsubsection*{Single-question confidence fine-tuning}
Although consistency scores are positively correlated with answer accuracy, they are not perfectly calibrated. In particular, consistency scores of 1 (indicating identical answers across all samples) are disproportionately frequent, reflecting the model’s tendency to be overconfident in certain cases (see Supplementary Figures A1 and A3 for results with GPT4.1-mini and Llama3.1-70B-instruct). To address this imbalance, in construction of the datasets for fine-tuning we applied a subsampling procedure to ensure a more uniform distribution of consistency scores in the training and test data. Specifically, we limited the frequency difference between the most common and the second-most common consistency score bins to no more than 20\%. This typically involved downsampling questions with perfect consistency (score = 1), resulting in an overrepresentation of questions with lower consistency and accuracy. This, in turn, reduced the average accuracy of the fine-tuning datasets (see Supplementary Tables A2  and A5).

The resulting datasets were split into training and test sets. For MMLU-PRO and GSM8K, we sampled 2,000 training and 1,000 test examples each. For TriviaQA, which is smaller, we used 800 training and 559 test questions. 

For the smaller TruthfulQA and MetaMedQA datasets which were not used to create training data sets, all questions were used for testing leading to 790 and 1373 instances respectively. For LegalBench, we sampled 600 training questions for training and 649 test questions although the training questions were not used for fine-tuning in our setup.

Training instances for absolute confidence assessment consist of question-answer pairs where the question is in the format answer is in the format: ``Question: [$Q$]'' where $Q$ is a multiple choice question with answer options, math question or a short-answer question. The answer is in the format ``The answer is [$x$] and my confidence score is [$\hat{c}$]'',  
where $x$ is the modal answer and  $\hat{c}$ is the target confidence score. 

The target confidence $\hat{c}$ for each question is defined as the empirical accuracy associated with its consistency score plus noise:
\[
\hat{c} = a(s) + \varepsilon
\]
Here, $s$ is the consistency score for the question, $a(s)$ is the empirically observed accuracy for all questions with that consistency score in the dataset, and $\varepsilon$ is a noise term sampled uniformly from the interval $[-0.05, 0.05]$. For instance, in MMLU-PRO, a consistency score of $8/10$ might correspond to an empirical accuracy of $0.35$, reflecting a correction for overconfidence. These mappings are derived from the calibration data shown in Supplementary Figures A1 and A3. This procedure differs from Lin et al. (2022), where the target confidence is computed at the subtask level (i.e., all questions from the same subtask share the same label). Here, the target confidence varies per-question depending on the per-instance consistency.

The noise discourages the model from learning exact mappings between the small set of possible consistency scores (only 11 discrete values with $N = 10$ samples) and its internal representations. Introducing this variability can be viewed as a form of target-value regularization intended to discourage the model from memorizing a small set of discrete confidence values and to encourage smoother confidence reporting across nearby values. We evaluate the effect of this design choice in an ablation experiment reported in the Results section.

\subsubsection*{Confidence comparison fine-tuning}
For the confidence comparison fine-tuning task, the goal was to fine-tune the model to determine, for any given pair of questions, which one it was more likely to answer correctly. To accomplish this, we created labeled training instances consisting of question-answer pairs. Each prompt followed the format shown in Supplementary Table 1, and the target response was structured as: ``The answer is [$y$],'' where $y\in\{Q1,Q2\}$ corresponds to the question with the higher consistency score.

To generate these training instances, we began with the same training and test set partitions used for the single-question confidence fine-tuning task. For the MMLU-PRO, GSM8K, and TriviaQA datasets, we sampled 2,000 random pairs of questions in which the two questions had different consistency scores. The order of the questions in each pair was randomized to ensure that the first and second questions were equally represented as the correct answer.

A similar procedure was applied to create the test sets. For MMLU-PRO, GSM8K, and TriviaQA, we generated 1,000 test instances per dataset using the same method. For the smaller TruthfulQA, MetaMedQA, and LegalBench datasets, which were not used for fine-tuning, we created 1,000 test pairs by randomly pairing questions. Each pair was required to have unequal consistency scores, and the correct answer (either Q1 or Q2) was balanced across the set to ensure equal representation.

\subsubsection*{Supervised Fine-tuning}
Both tasks were optimized using standard supervised objectives (cross-entropy between model predictions and target answers). For the GPT-4.1-mini model, fine-tuning was performed using the OpenAI API.  The learning rate multiplier was set to 2, and batch size was automatically selected by the API. The training set was run for 10 epochs. For the Llama3.1-70B model, fine-tuning was performed through the cloud service fireworks.ai using default parameters. The training set was run for 5 epochs. 

\subsection*{Performance Metrics}

We evaluate two key aspects of uncertainty communication: calibration and
metacognitive discrimination. Calibration assesses how well the model's stated
confidence reflects its empirical accuracy and is evaluated for single-question
responses, where the model reports a numeric confidence score for its own
answer. Metacognitive discrimination assesses whether the model's uncertainty
reports distinguish cases in which its answers are correct from cases in which
they are incorrect. We evaluate discrimination in two complementary ways. First,
we use AUC-based measures, which quantify whether correct answers tend to receive
higher confidence than incorrect answers, or, in the pairwise setting, whether
the model selects the item associated with higher reference confidence or actual
correctness. These AUC measures can be interpreted as c-statistics over
permissible pairs. Second, for the single-question setting, we compute an
answer-level information-theoretic measure,
$\mathrm{meta}\text{-}I_{r,\mathrm{ans}}^2$, which quantifies the proportion of
uncertainty about answer correctness that is removed by observing the model's
confidence bin. Together, these metrics separate calibration, rank-based
discrimination, and information-theoretic confidence informativeness.

\subsubsection*{Single-Question Confidence}
In this task, the model is prompted to answer a single question and produce a corresponding confidence score between 0 and 1.

\paragraph{Calibration.}

Calibration refers to whether the model’s stated confidence matches its empirical accuracy. For instance, if a model outputs a confidence of 0.9 across a set of predictions, those answers should be correct 90\% of the time. We quantify miscalibration using the \textit{Expected Calibration Error} (ECE), which measures the average absolute difference between predicted confidence and observed accuracy across $M$ bins. In all our experiments, we used $M=11$, with bins at equally spaced intervals of 0.1 (i.e., $[0-0.1)$, $[0.1-0.2)$, etc). Let $B_m$ be the set of predictions in bin $m$, $N$ the total number of predictions, and $\text{acc}(B_m)$ and $\text{conf}(B_m)$ denote the empirical accuracy and average confidence in bin $m$, respectively:

\begin{equation}
\text{ECE} = \sum_{m=1}^{M} \frac{|B_m|}{N} \left| \text{acc}(B_m) - \text{conf}(B_m) \right|
\end{equation}

\paragraph{Metacognitive Discrimination: AUC.}
Our first measure for metacognitive discrimination is based on AUC, which captures the model’s ability to rank correct responses above incorrect ones. In this context, AUC is the probability that a randomly selected correct answer receives a higher confidence score than a randomly selected incorrect answer \parencite{hanley1982meaning}. This is equivalent to the c-statistic:

\begin{equation}
\text{AUC} = \frac{C}{P}
\end{equation}

where $P$ is the number of permissible (correct vs. incorrect) pairs, and $C$ is the number of concordant pairs in which the correct answer has a higher confidence than the incorrect one. In research on human metacognition, this is sometimes called the AUC2, the area under the Type 2 ROC function that plots Type 2 hit rate vs. Type 2 false alarm rate \parencite{rahnev2025comprehensive}. 

Unlike accuracy or calibration, AUC focuses purely on ranking performance and is invariant to the absolute values of confidence scores. However, it can still be influenced by changes in task performance, particularly accuracy—due to shifts in task difficulty \parencite{fleming2014how, rahnev2025comprehensive}.  
In our experiments, this potential confound is minimized, as model accuracy remains largely stable across fine-tuning conditions, allowing us to interpret changes in AUC as reflecting improvements in metacognitive discrimination rather than changes in task difficulty.

\paragraph{Metacognitive Discrimination: Answer-Level Metainformation}

We supplemented the AUC-based discrimination analyses with an
information-theoretic measure of confidence informativeness. AUC asks whether
correct answers tend to receive higher confidence than incorrect answers. This
ranking interpretation is useful, but it does not measure how much uncertainty
about correctness is reduced by observing a confidence report. To capture this
quantity, we used an answer-level adaptation of Dayan's metacognitive
information framework \parencite{Dayan2023}.

Dayan's meta-$I$ measures the mutual information between response correctness
and confidence. In this framework, a confidence report is metacognitively
informative to the extent that it reduces uncertainty about whether the original
response was correct. Dayan also proposed normalized versions of meta-$I$,
including $\mathrm{meta}\text{-}I_r^2$, which divides meta-$I$ by the entropy of
correctness. This normalization accounts for the amount of uncertainty about
correctness that is available to be reduced: when responses are almost always
correct or almost always incorrect, there is little entropy in correctness, and
therefore little information for confidence to convey.

Subsequent work by \parencite{MeyenEtAl2025} extended the information-theoretic
approach by deriving accuracy-dependent bounds on transmitted information,
leading to the relative metainformation (RMI) measure. RMI is designed for
classifier responses that contain enough information to evaluate the relation
between the true task label, the predicted label, and confidence. In the
multiclass case, this requires information about how the classifier distributes
probability across possible labels, or enough data to estimate the distribution
of true labels conditional on the classifier's full response. The bounded RMI
measure therefore provides a stronger accuracy-dependent normalization than
Dayan's $\mathrm{meta}\text{-}I_r^2$, but it requires information that is not
available in our setting.

Our setting is different. The LLM provides a single answer and a single
confidence score for that answer. For open-ended questions, the model does not
provide confidence scores over all possible answers, and the set of possible
answers is not fixed in advance. Therefore, we do not compute multiclass RMI.
Instead, we compute an answer-level version of Dayan's
$\mathrm{meta}\text{-}I_r^2$ that asks how much the model's stated confidence
tells us about whether the answer it produced is correct.

For each question $i$, let $c_i$ denote the confidence score reported by the
model for its answer. We define a binary correctness variable
\begin{equation}
z_i =
\begin{cases}
1, & \text{if the model's answer to question } i \text{ is scored as correct},\\
0, & \text{otherwise}.
\end{cases}
\end{equation}
The same confidence bins used for the ECE calculation are used here. Let $B_m$
denote the set of trials whose confidence scores fall in bin $m$, let
$|B_m|/N$ denote the proportion of trials in that bin, and let
$\mathrm{acc}(B_m)$ denote the empirical accuracy in that bin. Thus, the
confidence score $c_i$ is used only to assign trial $i$ to a confidence bin; the
information calculation is based on the empirical relationship between
confidence bins and correctness.

We define answer-level meta-$I$ as the mutual information between correctness
and the confidence bin:
\begin{equation}
\mathrm{meta}\text{-}I_{\mathrm{ans}}
=
H(Z) - H(Z \mid B).
\end{equation}
Here, $Z$ denotes the binary correctness variable over trials, $H(Z)$ is the
entropy of answer correctness, and $H(Z \mid B)$ is the remaining uncertainty
about correctness after observing the confidence bin. In practice,
$H(Z \mid B)$ is computed from the empirical accuracy within each confidence
bin:
\begin{equation}
H(Z \mid B)
=
\sum_{m=1}^{M}
\frac{|B_m|}{N}
H_2\!\left(\mathrm{acc}(B_m)\right),
\end{equation}
where $H_2(\cdot)$ is the binary entropy function.

To express this information relative to the uncertainty available in answer
correctness, we normalize by the total entropy of correctness:
\begin{equation}
\mathrm{meta}\text{-}I_{r,\mathrm{ans}}^2
=
\frac{\mathrm{meta}\text{-}I_{\mathrm{ans}}}{H(Z)}.
\end{equation}
This measure is the answer-level analogue of Dayan's
$\mathrm{meta}\text{-}I_r^2$. It ranges from 0 to 1 when $H(Z)>0$. A value of 0
means that confidence bins provide no information about whether the model's
answer is correct. A value of 1 means that confidence bins completely determine
whether the answer is correct. The measure is undefined when all answers are
correct or all answers are incorrect, because there is then no uncertainty about
correctness to reduce.

The answer-level $\mathrm{meta}\text{-}I_{r,\mathrm{ans}}^2$ should be
interpreted as an information-theoretic measure of metacognitive discrimination
for the answer actually produced by the LLM. It measures the proportion of
uncertainty about answer correctness that is removed by observing the model's
reported confidence bin. This normalization accounts for one direct consequence
of accuracy differences: the amount of entropy in correctness changes with the
model's overall accuracy. However, the measure should not be interpreted as
fully accuracy-independent. Unlike the bounded RMI measure of
\parencite{MeyenEtAl2025}, it does not normalize by the tight lower and upper
information bounds available at a given Type 1 accuracy. We therefore report and
interpret answer-level $\mathrm{meta}\text{-}I_{r,\mathrm{ans}}^2$ alongside
accuracy, AUC, and ECE.

Finally, answer-level $\mathrm{meta}\text{-}I_{r,\mathrm{ans}}^2$ does not
measure calibration; calibration is assessed separately using ECE. It also does
not measure how much information the model has about the full set of possible
answers. Rather, it captures how informative the model's expressed confidence is
about the correctness of the specific answer it chose to give.

\subsubsection*{Confidence Comparison}

In the confidence comparison task, the model is presented with two questions and asked to indicate which one it is more confident in answering correctly. This task is inherently pairwise and does not require numerical confidence scores. We evaluate performance using two variants of AUC:

\paragraph{AUC based on Reference Consistency (AUCc).}

AUCc measures how often the model’s choice agrees with a reference ranking based on consistency scores (derived from sampling). For each permissible pair of questions with unequal consistency scores, we define the comparison as concordant if the model selects the question with the higher reference score:

\begin{equation}
\text{AUCc} = \frac{C_{\text{consistency}}}{P_{\text{consistency}}}
\end{equation}

\paragraph{AUC based on Answer Correctness (AUCa).}

AUCa measures how often the model selects the question it ultimately answers correctly. The permissible set includes all pairs where the model is correct on one question and incorrect on the other. A comparison is concordant if the model chooses the question it got right:

\begin{equation}
\text{AUCa} = \frac{C_{\text{accuracy}}}{P_{\text{accuracy}}}
\end{equation}

As with the single-question AUC, both AUCc and AUCa reflect the model’s ability to rank questions according to its internal uncertainty—whether derived from confidence scores or implicit judgments. These metrics share a common interpretation as c-statistics computed over appropriate pairings, reinforcing the idea that a single unified discrimination framework applies across both tasks.

\subsubsection*{Statistical Analysis}

We used paired bootstrap procedures to estimate uncertainty for changes in model performance before versus after fine-tuning. For each metric, the paired change was defined as the value after fine-tuning minus the value before fine-tuning. This pairing preserves the fact that the same test questions, or the same question pairs in the comparison task, were evaluated before and after fine-tuning.

For single-question confidence estimation, we evaluated calibration using Expected Calibration Error (ECE), discrimination using AUC, answer-level meta-information ($\mathrm{meta}\text{-}I_{r,\mathrm{ans}}^2$), and answer accuracy. For each bootstrap replicate, test questions were resampled with replacement within each dataset, keeping the before- and after-fine-tuning predictions for each sampled question paired. Metrics were then recomputed separately for each dataset and each fine-tuning condition. For single-question contrasts involving multiple datasets, paired changes were computed separately within each dataset, then averaged across datasets with weights proportional to the number of paired observations in each dataset

For the pairwise confidence comparison task, the resampling unit was the question pair. On each bootstrap replicate, question pairs were resampled with replacement within each dataset, again preserving the pairing between before- and after-fine-tuning outputs for the same question pair. We then recomputed AUC relative to reference consistency scores (AUCc) and AUC relative to answer correctness (AUCa). For multi-dataset contrasts, paired changes were computed separately within each dataset and then averaged across datasets with weights proportional to the number of paired observations in each dataset.

Confidence intervals were computed from the empirical bootstrap distribution using percentile 95\% intervals. Bootstrap-based two-sided tests were conducted by comparing the bootstrap distribution of paired changes against zero. We report two-sided $p$ values for planned contrasts in the Results section; values smaller than the resolution of the bootstrap procedure with 1,000 resamples were reported as $p = .001$. 

\section*{Statements \& Declarations}
\subsection*{Data Availability}
All question data sets, training data, and responses produced by the LLMs are available from the following OSF repository: \url{https://osf.io/k32wa/overview?view_only=f26ef0708a5d49ab8312beb8adfe3314}

\subsection*{Code Availability}
The code used for analyzing the LLM responses and recreating the tables and figures in the paper is available from the following OSF repository: \url{https://osf.io/k32wa/overview?view_only=f26ef0708a5d49ab8312beb8adfe3314} 

\subsection*{Funding Statement}
This research has been supported by NSF grant IIS-2505006.

\subsection*{Author Contributions}
M.S., C.B., and P.S. developed the general framework. M.S. performed the analyses. M.S., C.B., and P.S. wrote the main manuscript text. All authors reviewed the manuscript.

\subsection*{Acknowledgments}
Not applicable

\subsection*{Competing Interests}
The authors declare no competing interests.



\printbibliography

\appendix
\newpage
\renewcommand{\figurename}{Supplementary Figure}
\renewcommand{\tablename}{Supplementary Table}

\renewcommand{\thefigure}{\arabic{figure}}
\setcounter{figure}{0}

\renewcommand{\thetable}{\arabic{table}}
\setcounter{table}{0}

\setcounter{page}{1}

\section*{Supplementary Materials}

\section*{Prompts}

\begin{table}[H]
\small
  \caption{Prompts to elicit answers and confidence. For the single-question confidence task, in addition to the prompt, a system prompt was used as well in the form ``When answering a question, provide the answer and a confidence score between 0 and 1 for the answer''.}
  \label{tab:exampleprompts}
  \centering
  \begin{tabular}{p{2.5cm}p{6cm}p{6cm}}
    \toprule
  Data Set & Single-question Confidence Prompt & Confidence Comparison Prompt\\
    \midrule
   MMLU-PRO, TruthfulQA, MetaMedQA & Question: [QUESTION]\newline
   Options:\newline A: [RESPONSE OPTION A]\newline B: [RESPONSE OPTION B]\newline ...\newline Answer:\newline
     Confidence score (0-1): & Determine for which of the two following questions, your confidence score is higher\newline \newline Q1: [QUESTION 1]\newline A: [RESPONSE OPTION A]\newline B: [RESPONSE OPTION B]\newline ... \newline \newline Q2: [QUESTION 2]\newline A: [RESPONSE OPTION A]\newline B: [RESPONSE OPTION B]\newline ... \newline Is your confidence in answering correctly higher for question Q1 or question Q2? What are the answers to Q1 and Q2?\\
\addlinespace     
   GSM8K & Question: [QUESTION]\newline Answer:\newline
     Confidence score (0-1): & Determine for which of the two following questions, your confidence score is higher\newline \newline Q1: [QUESTION 1]\newline  Q2: [QUESTION 2]\newline  \newline Is your confidence in answering correctly higher for question Q1 or question Q2? What are the answers to Q1 and Q2?\\
\addlinespace     
   TriviaQA & For the following problem, provide a short answer with just a few words.\newline Question: [QUESTION]\newline Answer:\newline Confidence score (0-1):\newline & Determine for which of the two following questions, your confidence score is higher\newline \newline Q1: [QUESTION 1]\newline  Q2: [QUESTION 2]\newline  \newline Is your confidence in answering correctly higher for question Q1 or question Q2? What are the answers to Q1 and Q2?\\
  \bottomrule
\end{tabular}
\end{table}


\section*{Additional Results for GPT-4.1-mini}

\begin{figure}[H]
\centering
\includegraphics[width=0.85\textwidth]{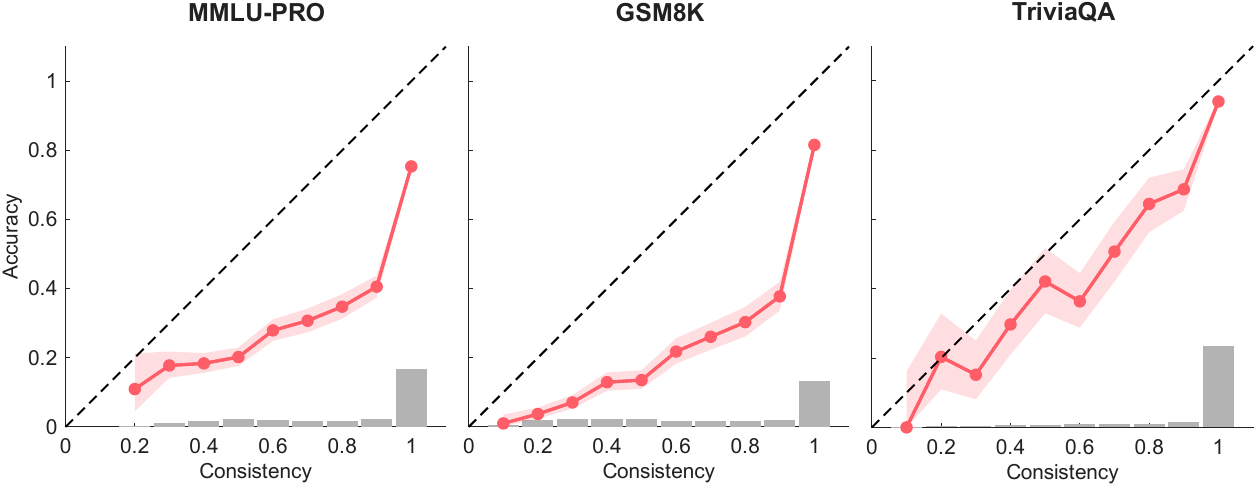}
\includegraphics[width=0.85\textwidth]{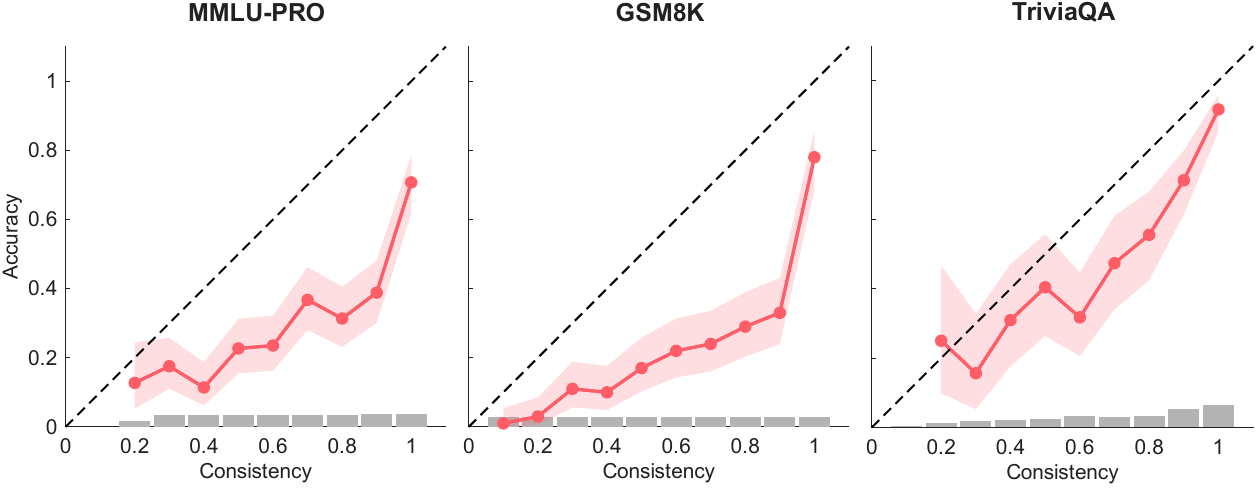}
\caption{Calibration results for consistency scores across task domains for GPT4.1 Mini. Results are based on the full data set before subsampling (top row) and the test set after subsampling (bottom row). The histograms at the bottom of each plot show the proportion of observations in each confidence bin (values are scaled by 30\% for visual clarity).}
\label{fig:AppendixConsistencyCalibration}
\end{figure}

\begin{table}[h]
  \caption{Discrimination (AUC), Calibration (ECE), and accuracy for consistency scores across task domains.}
  \centering
  \begin{tabular}{lllllll}
    \toprule
  Task Domain &  Set & N & Accuracy & AUC & ECE\\
    \midrule
  MMLU-PRO & Full data set & 12032 & 0.54 & 0.75 & 0.29 \\
           & Test set of subsampled data set & 1000 & 0.31 & 0.71 & 0.32 \\ 
  \addlinespace
  GSM8K    & Full data set & 8737 & 0.46 & 0.86 & 0.27 \\
           & Test set of subsampled data set&1000 & 0.23 & 0.80 & 0.33 \\
  \addlinespace
  TriviaQA & Full data set & 4950 & 0.84 & 0.81 & 0.08 \\
           & Test set of subsampled data set & 559& 0.56 & 0.79 & 0.16 \\
  \bottomrule
\end{tabular}
\label{tab:AppendixConsistencyFull}
\end{table}

\begin{table}[h]
  \caption{Single-question confidence results before and after fine-tuning GPT-4.1-mini across different types of generalization tests. LLMs were finetuned on the single-question confidence task (S), the confidence comparison task (C), or the combination of both tasks (C+S). AMI represents answer-level meta-information ($\mathrm{meta}\text{-}I_{r,\mathrm{ans}}^2$).}
  \centering
  \small
  \begin{tabular}{
p{0.6cm}
lll@{\hspace{0.28em}}
l@{\hspace{0.22em}}l
l@{\hspace{0.38em}}
l@{\hspace{0.22em}}l
l@{\hspace{0.38em}}
l@{\hspace{0.22em}}l
l@{\hspace{0.38em}}
l@{\hspace{0.22em}}l
}
    \toprule
    & \multicolumn{2}{c}{Domain} && \multicolumn{2}{c}{AUC}   
                                 && \multicolumn{2}{c}{AMI}   
                                 && \multicolumn{2}{c}{ECE}   
                                 && \multicolumn{2}{c}{Accuracy} \\
    \cline{2-3} \cline{5-6} \cline{8-9}  \cline{11-12}  \cline{14-15}
  Type & Test Domain & Training Domain && Before & After      && Before & After     && Before & After & & Before & After\\
    \midrule
\multirow{1}{*}{Within Domain}\\
&             MMLU-PRO &               M (S) && 0.52 & 0.68  &&     0.00 & 0.08   &&     0.61 & 0.05      && 0.30 & 0.33\\
&                GSM8K &               G (S) && 0.53 & 0.76  &&     0.00 & 0.16   &&     0.75 & 0.04      && 0.21 & 0.21\\
&             TriviaQA &               T (S) && 0.75 & 0.83  &&     0.05 & 0.27   &&     0.41 & 0.08      && 0.49 & 0.50\\
&             MMLU-PRO &           M+G+T (S) && 0.52 & 0.67  &&     0.00 & 0.07   &&     0.61 & 0.04      && 0.30 & 0.33\\
&                GSM8K &           M+G+T (S) && 0.53 & 0.77  &&     0.00 & 0.17   &&     0.75 & 0.06      && 0.21 & 0.20\\
&             TriviaQA &           M+G+T (S) && 0.75 & 0.86  &&     0.05 & 0.30   &&     0.42 & 0.08      && 0.49 & 0.48\\
\addlinespace
\multirow{1}{*}{Across Domain}\\
&           TruthfulQA &           M+G+T (S) && 0.67 & 0.73  &&     0.08 & 0.19   &&     0.10 & 0.18      && 0.81 & 0.78\\
&            MetaMedQA &           M+G+T (S) && 0.71 & 0.75  &&     0.10 & 0.13   &&     0.23 & 0.18      && 0.69 & 0.68\\
&           LegalBench &           M+G+T (S) && 0.58 & 0.63  &&     0.01 & 0.07   &&     0.33 & 0.20      && 0.58 & 0.62\\
&             MMLU-PRO &               G (S) && 0.52 & 0.62  &&     0.00 & 0.04   &&     0.61 & 0.08      && 0.30 & 0.26\\
&             MMLU-PRO &               T (S) && 0.53 & 0.55  &&     0.00 & 0.02   &&     0.61 & 0.20      && 0.30 & 0.26\\
&                GSM8K &               M (S) && 0.53 & 0.61  &&     0.00 & 0.04   &&     0.75 & 0.16      && 0.21 & 0.19\\
&                GSM8K &               T (S) && 0.53 & 0.63  &&     0.00 & 0.06   &&     0.75 & 0.38      && 0.21 & 0.06\\
&             TriviaQA &               M (S) && 0.75 & 0.78  &&     0.05 & 0.20   &&     0.41 & 0.04      && 0.49 & 0.49\\
&             TriviaQA &               G (S) && 0.75 & 0.71  &&     0.05 & 0.11   &&     0.41 & 0.10      && 0.49 & 0.51\\
\addlinespace
\multirow{1}{*}{Across Tasks}\\
&             MMLU-PRO &           M+G+T (C) && 0.52 & 0.60  &&     0.00 & 0.01   &&     0.61 & 0.60      && 0.30 & 0.30\\
&                GSM8K &           M+G+T (C) && 0.53 & 0.55  &&     0.00 & 0.00   &&     0.75 & 0.74      && 0.21 & 0.21\\
&             TriviaQA &           M+G+T (C) && 0.75 & 0.78  &&     0.05 & 0.10   &&     0.41 & 0.43      && 0.49 & 0.49\\
&           TruthfulQA &           M+G+T (C) && 0.67 & 0.63  &&     0.07 & 0.05   &&     0.10 & 0.13      && 0.81 & 0.78\\
&            MetaMedQA &           M+G+T (C) && 0.71 & 0.69  &&     0.10 & 0.08   &&     0.23 & 0.27      && 0.69 & 0.67\\
&           LegalBench &           M+G+T (C) && 0.58 & 0.58  &&     0.01 & 0.03   &&     0.33 & 0.27      && 0.58 & 0.66\\
\addlinespace
\multirow{1}{*}{Combined Tasks}\\
&             MMLU-PRO &           M+G+T (C+S) && 0.52 & 0.67  &&     0.00 & 0.07   &&     0.61 & 0.03      && 0.30 & 0.31\\
&                GSM8K &           M+G+T (C+S) && 0.53 & 0.76  &&     0.00 & 0.15   &&     0.75 & 0.06      && 0.21 & 0.22\\
&             TriviaQA &           M+G+T (C+S) && 0.75 & 0.85  &&     0.05 & 0.32   &&     0.41 & 0.11      && 0.49 & 0.46\\
&           TruthfulQA &           M+G+T (C+S) && 0.67 & 0.76  &&     0.07 & 0.14   &&     0.10 & 0.13      && 0.81 & 0.75\\
&            MetaMedQA &           M+G+T (C+S) && 0.71 & 0.79  &&     0.10 & 0.19   &&     0.23 & 0.15      && 0.69 & 0.67\\
&           LegalBench &           M+G+T (C+S) && 0.58 & 0.66  &&     0.01 & 0.07   &&     0.33 & 0.15      && 0.58 & 0.59\\
  \bottomrule
\end{tabular}
\label{tab:absconfallresults1}
\end{table}

\begin{figure}
\includegraphics[width=\textwidth]{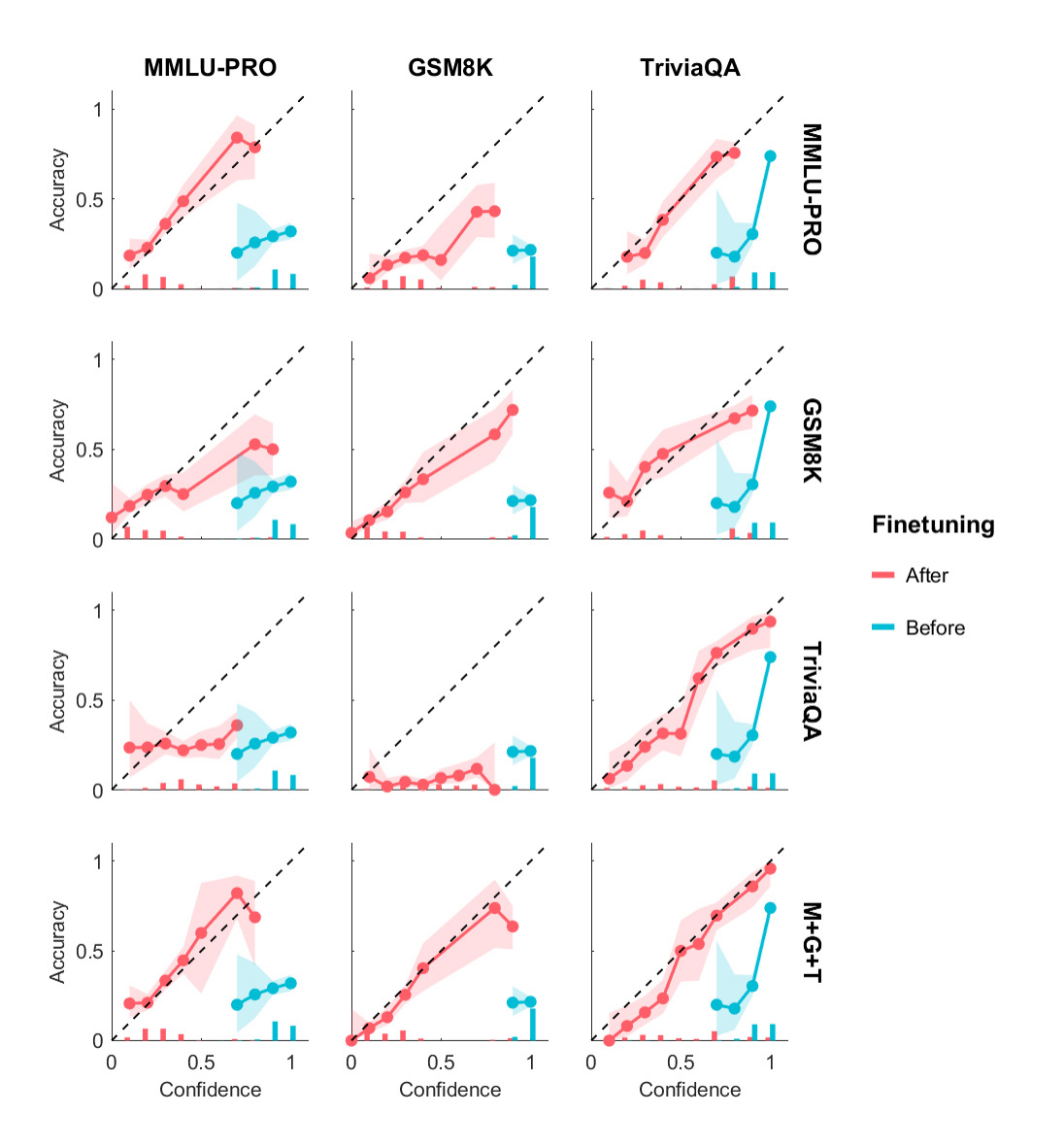}
\caption{Calibration results for baseline and finetuned models for the single-question confidence task for GPT4.1 Mini. Rows correspond to data sets used for training. Columns correspond to data sets used for testing. Note that no questions used for finetuning models ever appear in the test data. The histograms at the bottom of each plot show the proportion of observations in each confidence bin (values are scaled by 30\% for visual clarity).}
\label{fig:fullconfresults3}
\end{figure}

\begin{table}[h]
  \caption{Confidence comparison results before and after fine-tuning of GPT-4.1-mini.  LLMs were finetuned on the single-question confidence task (S), the confidence comparison task (C), or the combination of both tasks (C+S). A row index (\#) is added for ease of referencing.}
  \small
  \centering
  \begin{tabular}{lllllllllr}
    \toprule
    & \multicolumn{2}{c}{Domain} & & \multicolumn{2}{c}{AUCc} & & \multicolumn{2}{c}{AUCa}   \\
   \cline{2-3}  \cline{5-6} \cline{8-9}
  Type & Test & Training && Before & After & & Before & After & \#\\
    \midrule
Within Domain\\
&             MMLU-PRO &        MMLU-PRO (C) && 0.57 & 0.72 && 0.55 & 0.64 & 1\\
&                GSM8K &           GSM8K (C) && 0.61 & 0.75 && 0.75 & 0.74 & 2\\
&             TriviaQA &        TriviaQA (C) && 0.68 & 0.74 && 0.78 & 0.81 & 3\\
    \addlinespace
&             MMLU-PRO &           M+G+T (C) && 0.57 & 0.72 && 0.55 & 0.65 & 4\\
&                GSM8K &           M+G+T (C) && 0.61 & 0.74 && 0.75 & 0.73 & 5\\
&             TriviaQA &           M+G+T (C) && 0.68 & 0.73 && 0.78 & 0.79 & 6\\
    \addlinespace
Across Domain\\
&           TruthfulQA &           M+G+T (C) && 0.59 & 0.68 && 0.47 & 0.56 & 7\\
&            MetaMedQA &           M+G+T (C) && 0.67 & 0.75 && 0.66 & 0.80 & 8\\
&           LegalBench &           M+G+T (C) && 0.55 & 0.58 && 0.70 & 0.66 & 9\\

\addlinespace
&             MMLU-PRO &           GSM8K (C) && 0.57 & 0.59 && 0.55 & 0.61 & 10\\
&             MMLU-PRO &        TriviaQA (C) && 0.57 & 0.59 && 0.55 & 0.56 & 11\\
&                GSM8K &        MMLU-PRO (C) && 0.61 & 0.62 && 0.75 & 0.70 & 12\\
&                GSM8K &        TriviaQA (C) && 0.61 & 0.63 && 0.75 & 0.73 & 13\\
&             TriviaQA &        MMLU-PRO (C) && 0.68 & 0.67 && 0.78 & 0.73 & 14\\
&             TriviaQA &           GSM8K (C) && 0.68 & 0.65 && 0.78 & 0.69 & 15\\

\addlinespace
Across Tasks\\
&             MMLU-PRO &           M+G+T (S) && 0.57 & 0.57 && 0.55 & 0.54 & 16\\
&                GSM8K &           M+G+T (S) && 0.61 & 0.66 && 0.75 & 0.68 & 17\\
&             TriviaQA &           M+G+T (S) && 0.68 & 0.69 && 0.78 & 0.79 & 18\\
&           TruthfulQA &           M+G+T (S) && 0.59 & 0.57 && 0.47 & 0.44 & 19\\
&            MetaMedQA &           M+G+T (S) && 0.67 & 0.67 && 0.66 & 0.74 & 20\\
&           LegalBench &           M+G+T (S) && 0.55 & 0.50 && 0.70 & 0.74 & 21\\

\addlinespace
Combined Tasks\\
&             MMLU-PRO &           M+G+T (C+S) && 0.57 & 0.72 && 0.55 & 0.63 & 22\\
&                GSM8K &           M+G+T (C+S) && 0.61 & 0.76 && 0.75 & 0.79 & 23\\
&             TriviaQA &           M+G+T (C+S) && 0.68 & 0.72 && 0.78 & 0.80 & 24\\
&           TruthfulQA &           M+G+T (C+S) && 0.59 & 0.71 && 0.47 & 0.58 & 25\\
&            MetaMedQA &           M+G+T (C+S) && 0.67 & 0.78 && 0.66 & 0.78 & 26\\
&           LegalBench &           M+G+T (C+S) && 0.55 & 0.71 && 0.70 & 0.73 & 27\\

  \bottomrule
\end{tabular}
\label{tab:relconf1fullresults}
\end{table}

\begin{figure}[H]
\centering
\includegraphics[width=0.85\textwidth]{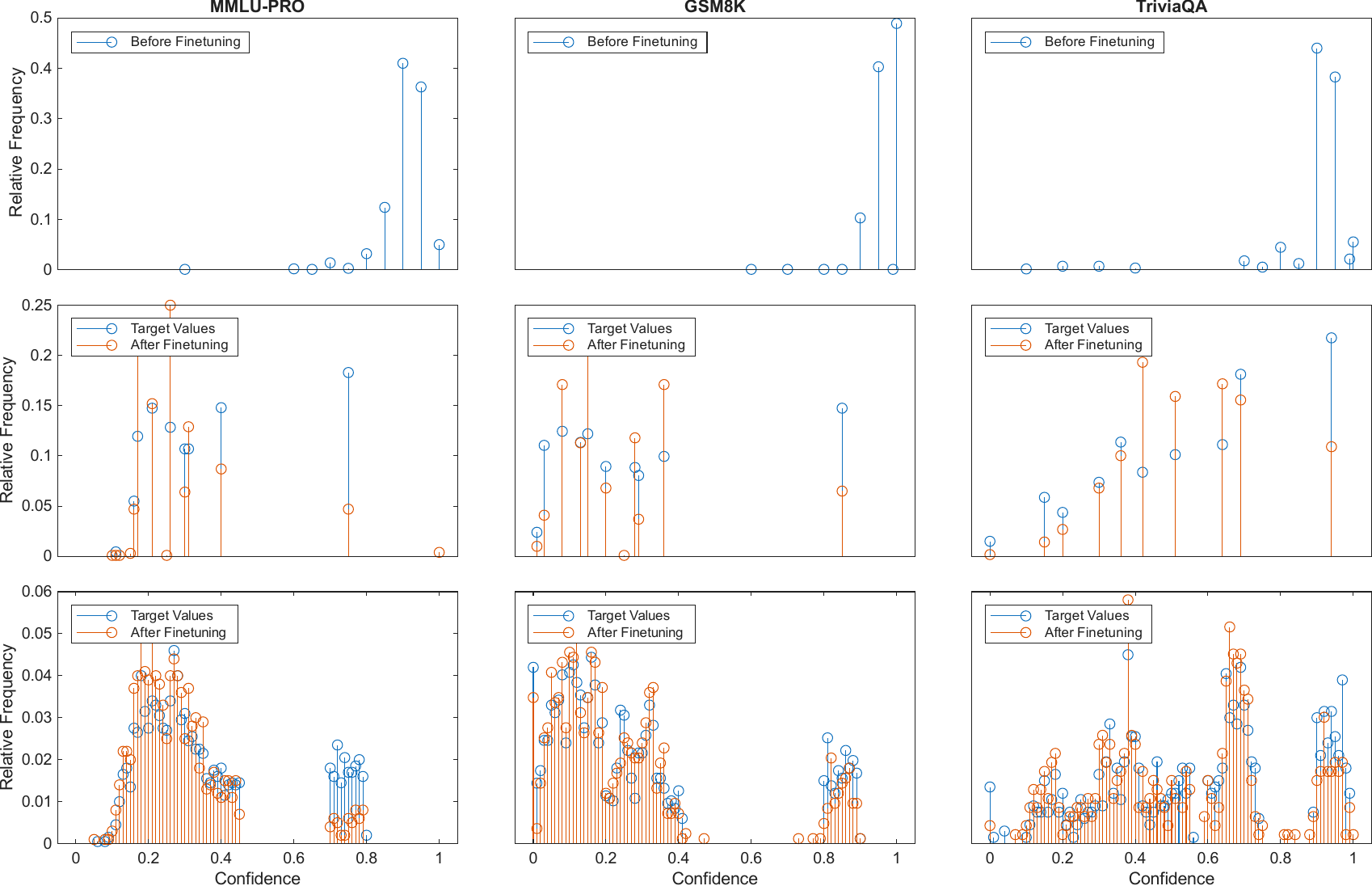}
\caption{Effect of target-confidence noise on confidence values after fine-tuning GPT-4.1-mini. Results are shown for within-domain fine-tuning on MMLU-PRO, GSM8K, and TriviaQA, corresponding to Experiments 1–3 in Table 1. The top row shows the confidence values produced before fine-tuning. The middle row compares the target confidence values used during fine-tuning with the confidence values produced after fine-tuning when no noise was added to the targets. The bottom row shows the same comparison when uniform noise, $\epsilon \sim U(-0.05, 0.05)$, was added to the target confidence values. Adding noise increases the number of distinct confidence values produced after fine-tuning, indicating that noise helps reduce clustering around a small set of discrete target values.}
\label{fig:AppendixConfidenceNoise}
\end{figure}

\FloatBarrier
\section*{Additional Results for Llama3.1-70B}

\begin{figure}[H]
\includegraphics[width=\textwidth]{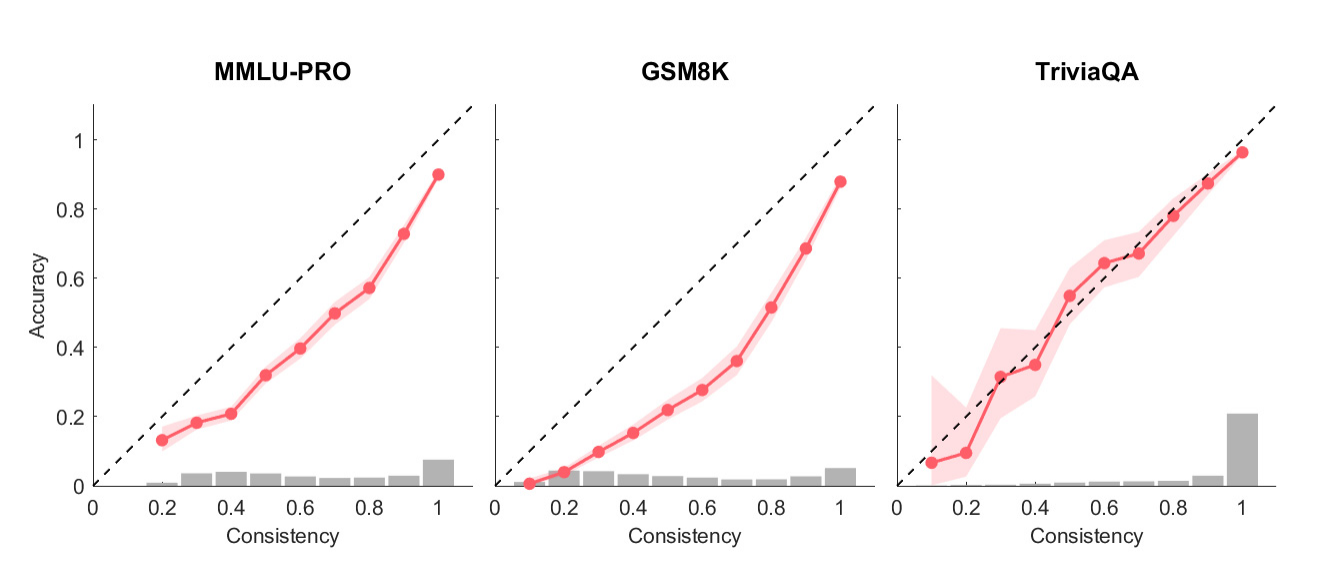}
\includegraphics[width=\textwidth]{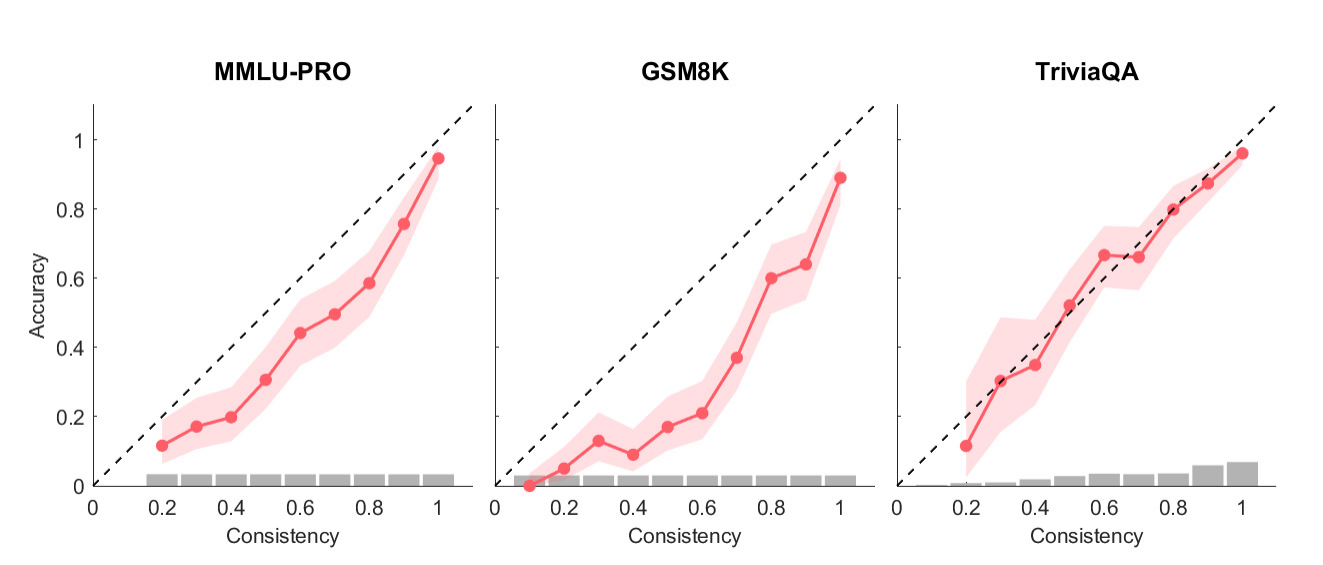}
\caption{Calibration results for consistency scores across task domains for Llama3.1-70B. Results are based on the full data set before subsampling (top row) and the test set after subsampling (bottom row). The histograms at the bottom of each plot show the proportion of observations in each confidence bin (values are scaled by 30\% for visual clarity)}
\label{fig:AppendixConsistencyCalibrationLlama}
\end{figure}

\begin{table}[h]
  \caption{Discrimination (AUC), calibration (ECE), and accuracy for consistency scores across task domains.}
  \centering
  \begin{tabular}{llllll}
    \toprule
  Task Domain &  Set & N & Accuracy & AUC & ECE\\
    \midrule
  MMLU-PRO & Full data set     & 12032 & 0.51 & 0.81 & 0.16 \\
           & Test set of subsampled data set & 1000 & 0.45 & 0.80 & 0.15 \\ 
  \addlinespace
  GSM8K    & Full data set     & 8737 & 0.35 & 0.87 & 0.21 \\
           & Test set of subsampled data set &1000 & 0.32 & 0.86 & 0.23 \\
  \addlinespace
  TriviaQA & Full data set     & 4950 & 0.88 & 0.82 & 0.04 \\
           & Test set of subsampled data set & 1000 & 0.72 & 0.80 & 0.03 \\
  \bottomrule
\end{tabular}
\label{tab:AppendixConsistencyFullLlama}
\end{table}



\begin{figure}
\includegraphics[width=\textwidth]{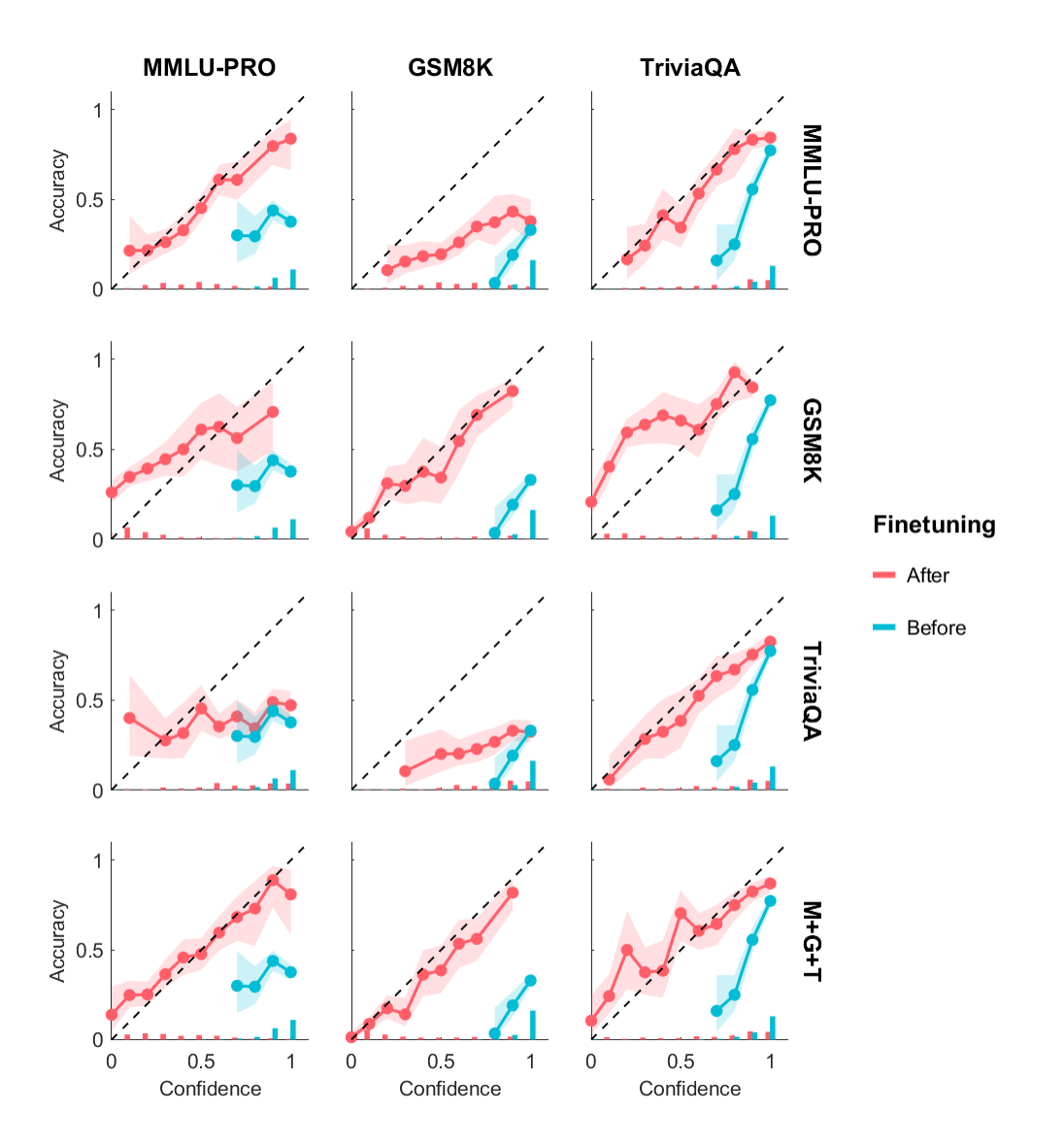}
\caption{Calibration results for baseline and finetuned models for the single-question confidence task for Llama3.1-70B. Rows correspond to data sets used for training. Columns correspond to data sets used for testing. Note that no questions used for finetuning models ever appear in the test data. The histograms at the bottom of each plot show the proportion of observations in each confidence bin (values are scaled by 30\% for visual clarity).}
\label{fig:fullconfresults11}
\end{figure}

\begin{table}[h]
\caption{Single-question confidence results before and after fine-tuning Llama3.1-70B across different types of generalization tests. LLMs were finetuned on the single-question confidence task (S), the confidence comparison task (C), or the combination of both tasks (C+S). AMI represents answer-level meta-information ($\mathrm{meta}\text{-}I_{r,\mathrm{ans}}^2$).}
  \centering
  \small
  \begin{tabular}{
p{0.6cm}
lll@{\hspace{0.28em}}
l@{\hspace{0.22em}}l
l@{\hspace{0.38em}}
l@{\hspace{0.22em}}l
l@{\hspace{0.38em}}
l@{\hspace{0.22em}}l
l@{\hspace{0.38em}}
l@{\hspace{0.22em}}l
}
    \toprule
    & \multicolumn{2}{c}{Domain} && \multicolumn{2}{c}{AUC}   
                                 && \multicolumn{2}{c}{AMI}   
                                 && \multicolumn{2}{c}{ECE}   
                                 && \multicolumn{2}{c}{Accuracy} \\
    \cline{2-3} \cline{5-6} \cline{8-9}  \cline{11-12}  \cline{14-15}
  Type & Test Domain & Training Domain && Before & After      && Before & After     && Before & After & & Before & After\\
    \midrule
\multirow{1}{*}{Within Domain}\\
&             MMLU-PRO &               M (S) && 0.51 & 0.73  &&     0.01 & 0.14   &&     0.54 & 0.06      && 0.38 & 0.45\\
&                GSM8K &               G (S) && 0.62 & 0.83  &&     0.02 & 0.27   &&     0.66 & 0.04      && 0.29 & 0.28\\
&             TriviaQA &               T (S) && 0.73 & 0.74  &&     0.13 & 0.16   &&     0.27 & 0.11      && 0.64 & 0.64\\
&             MMLU-PRO &           M+G+T (S) && 0.51 & 0.71  &&     0.01 & 0.10   &&     0.54 & 0.06      && 0.38 & 0.43\\
&                GSM8K &           M+G+T (S) && 0.62 & 0.84  &&     0.02 & 0.27   &&     0.66 & 0.05      && 0.30 & 0.26\\
&             TriviaQA &           M+G+T (S) && 0.73 & 0.76  &&     0.13 & 0.17   &&     0.27 & 0.08      && 0.64 & 0.67\\
\addlinespace
\multirow{1}{*}{Across Domain}\\
&           TruthfulQA &           M+G+T (S) && 0.66 & 0.67  &&     0.06 & 0.06   &&     0.08 & 0.18      && 0.80 & 0.82\\
&            MetaMedQA &           M+G+T (S) && 0.54 & 0.71  &&     0.02 & 0.10   &&     0.27 & 0.09      && 0.62 & 0.62\\
&           LegalBench &           M+G+T (S) && 0.60 & 0.64  &&     0.03 & 0.05   &&     0.34 & 0.08      && 0.56 & 0.52\\
&             MMLU-PRO &               G (S) && 0.51 & 0.63  &&     0.01 & 0.04   &&     0.54 & 0.20      && 0.38 & 0.39\\
&             MMLU-PRO &               T (S) && 0.51 & 0.56  &&     0.01 & 0.01   &&     0.54 & 0.32      && 0.38 & 0.40\\
&                GSM8K &               M (S) && 0.62 & 0.65  &&     0.02 & 0.04   &&     0.66 & 0.33      && 0.29 & 0.27\\
&                GSM8K &               T (S) && 0.62 & 0.60  &&     0.02 & 0.03   &&     0.66 & 0.51      && 0.30 & 0.27\\
&             TriviaQA &               M (S) && 0.73 & 0.76  &&     0.13 & 0.17   &&     0.27 & 0.08      && 0.64 & 0.67\\
&             TriviaQA &               G (S) && 0.73 & 0.74  &&     0.13 & 0.13   &&     0.27 & 0.19      && 0.64 & 0.63\\
\addlinespace
\multirow{1}{*}{Across Tasks}\\
&             MMLU-PRO &           M+G+T (C) && 0.51 & 0.55  &&     0.01 & 0.02   &&     0.54 & 0.48      && 0.38 & 0.43\\
&                GSM8K &           M+G+T (C) && 0.62 & 0.59  &&     0.02 & 0.04   &&     0.66 & 0.66      && 0.30 & 0.28\\
&             TriviaQA &           M+G+T (C) && 0.73 & 0.67  &&     0.13 & 0.11   &&     0.27 & 0.28      && 0.64 & 0.65\\
&           TruthfulQA &           M+G+T (C) && 0.66 & 0.61  &&     0.06 & 0.05   &&     0.08 & 0.15      && 0.81 & 0.78\\
&            MetaMedQA &           M+G+T (C) && 0.54 & 0.58  &&     0.02 & 0.05   &&     0.27 & 0.27      && 0.62 & 0.63\\
&           LegalBench &           M+G+T (C) && 0.60 & 0.62  &&     0.03 & 0.04   &&     0.34 & 0.32      && 0.56 & 0.60\\
\addlinespace
\multirow{1}{*}{Combined Tasks}\\
&             MMLU-PRO &           M+G+T (C+S) && 0.51 & 0.76  &&     0.01 & 0.16   &&     0.54 & 0.02      && 0.38 & 0.47\\
&                GSM8K &           M+G+T (C+S) && 0.62 & 0.84  &&     0.02 & 0.28   &&     0.66 & 0.08      && 0.29 & 0.27\\
&             TriviaQA &           M+G+T (C+S) && 0.73 & 0.77  &&     0.13 & 0.16   &&     0.27 & 0.07      && 0.64 & 0.67\\
&           TruthfulQA &           M+G+T (C+S) && 0.66 & 0.63  &&     0.06 & 0.03   &&     0.08 & 0.18      && 0.81 & 0.84\\
&            MetaMedQA &           M+G+T (C+S) && 0.54 & 0.75  &&     0.02 & 0.14   &&     0.27 & 0.05      && 0.62 & 0.63\\
&           LegalBench &           M+G+T (C+S) && 0.60 & 0.71  &&     0.03 & 0.10   &&     0.34 & 0.03      && 0.56 & 0.53\\
\bottomrule
\end{tabular}
\label{tab:absconfallresults1llama}
\end{table}

\begin{table}[h]
  \caption{Confidence comparison results before and after fine-tuning of Llama3.1-70B.  LLMs were finetuned on the single-question confidence task (S), the confidence comparison task (C), or the combination of both tasks (C+S). A row index (\#) is added for ease of referencing.}
  \small
  \centering
  \begin{tabular}{lllllllllr}
    \toprule
    & \multicolumn{2}{c}{Domain} & & \multicolumn{2}{c}{AUCc} & & \multicolumn{2}{c}{AUCa}   \\
   \cline{2-3}  \cline{5-6} \cline{8-9}
  Type & Test & Training && Before & After & & Before & After & \#\\
    \midrule
Within Domain\\
&             MMLU-PRO &        MMLU-PRO (C) && 0.55 & 0.77 && 0.54 & 0.69 & 1\\
&                GSM8K &           GSM8K (C) && 0.61 & 0.71 && 0.58 & 0.74 & 2\\
&             TriviaQA &        TriviaQA (C) && 0.56 & 0.62 && 0.67 & 0.66 & 3\\
    \addlinespace
&             MMLU-PRO &           M+G+T (C) && 0.55 & 0.77 && 0.54 & 0.66 & 4\\
&                GSM8K &           M+G+T (C) && 0.61 & 0.71 && 0.58 & 0.75 & 5\\
&             TriviaQA &           M+G+T (C) && 0.56 & 0.68 && 0.67 & 0.71 & 6\\
    \addlinespace
Across Domain\\
&           TruthfulQA &           M+G+T (C) && 0.53 & 0.69 && 0.62 & 0.64 & 7\\
&            MetaMedQA &           M+G+T (C) && 0.58 & 0.74 && 0.55 & 0.72 & 8\\
&           LegalBench &           M+G+T (C) && 0.59 & 0.65 && 0.58 & 0.67 & 9\\

\addlinespace
&             MMLU-PRO &           GSM8K (C) && 0.55 & 0.61 && 0.54 & 0.59 & 10\\
&             MMLU-PRO &        TriviaQA (C) && 0.55 & 0.54 && 0.54 & 0.57 & 11\\
&                GSM8K &        MMLU-PRO (C) && 0.61 & 0.61 && 0.58 & 0.60 & 12\\
&                GSM8K &        TriviaQA (C) && 0.61 & 0.54 && 0.58 & 0.54 & 13\\
&             TriviaQA &        MMLU-PRO (C) && 0.56 & 0.62 && 0.67 & 0.67 & 14\\
&             TriviaQA &           GSM8K (C) && 0.56 & 0.56 && 0.67 & 0.65 & 15\\

\addlinespace
Across Tasks\\
&             MMLU-PRO &           M+G+T (S) && 0.55 & 0.56 && 0.54 & 0.54 & 16\\
&                GSM8K &           M+G+T (S) && 0.61 & 0.60 && 0.58 & 0.56 & 17\\
&             TriviaQA &           M+G+T (S) && 0.56 & 0.56 && 0.67 & 0.67 & 18\\
&           TruthfulQA &           M+G+T (S) && 0.53 & 0.54 && 0.62 & 0.58 & 19\\
&            MetaMedQA &           M+G+T (S) && 0.58 & 0.57 && 0.55 & 0.55 & 20\\
&           LegalBench &           M+G+T (S) && 0.59 & 0.58 && 0.58 & 0.60 & 21\\

\addlinespace
Combined Tasks\\
&             MMLU-PRO &           M+G+T (C+S) && 0.55 & 0.76 && 0.54 & 0.67 & 22\\
&                GSM8K &           M+G+T (C+S) && 0.61 & 0.71 && 0.58 & 0.73 & 23\\
&             TriviaQA &           M+G+T (C+S) && 0.56 & 0.63 && 0.67 & 0.66 & 24\\
&           TruthfulQA &           M+G+T (C+S) && 0.53 & 0.65 && 0.62 & 0.61 & 25\\
&            MetaMedQA &           M+G+T (C+S) && 0.58 & 0.72 && 0.55 & 0.72 & 26\\
&           LegalBench &           M+G+T (C+S) && 0.59 & 0.61 && 0.58 & 0.63 & 27\\

  \bottomrule
\end{tabular}
\label{tab:relconf1fullresults}
\end{table}

\FloatBarrier
\section*{Additional Results for Qwen3 VL 30B A3B Instruct}

\begin{figure}[H]
\includegraphics[width=\textwidth]{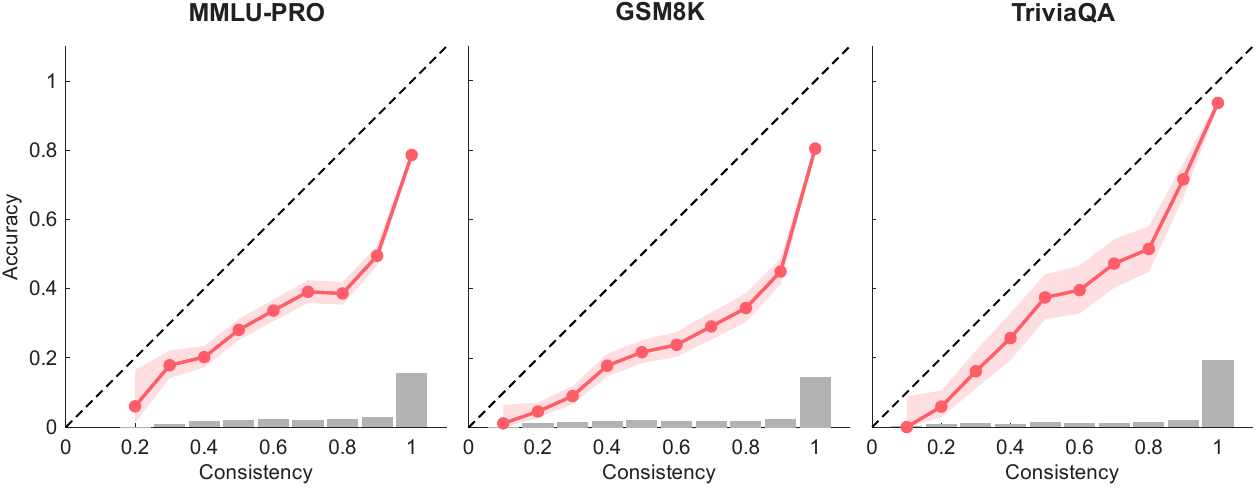}
\includegraphics[width=\textwidth]{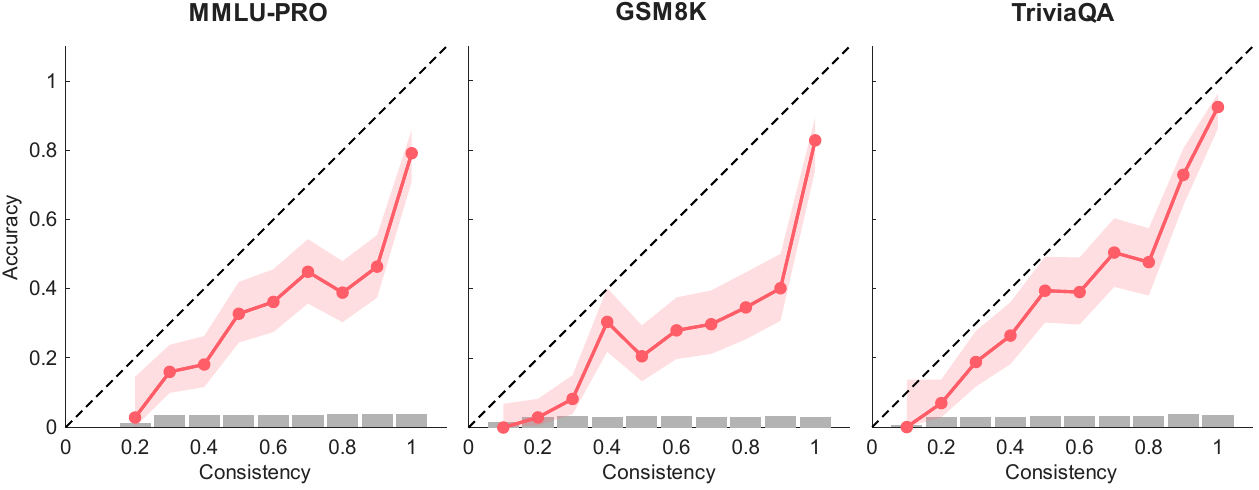}
\caption{Calibration results for consistency scores across task domains for Qwen3 VL 30B A3B Instruct. Results are based on the full data set before subsampling (top row) and the test set after subsampling (bottom row). The histograms at the bottom of each plot show the proportion of observations in each confidence bin (values are scaled by 30\% for visual clarity)}
\label{fig:AppendixConsistencyCalibrationQwen3}
\end{figure}

\begin{table}[h]
  \caption{Discrimination (AUC), calibration (ECE), and accuracy for consistency scores across task domains.}
  \centering
  \begin{tabular}{llllll}
    \toprule
  Task Domain &  Set & N & Accuracy & AUC & ECE\\
    \midrule
  MMLU-PRO & Full data set     & 12032 & 0.58 & 0.75 & 0.25 \\
           & Test set of subsampled data set & 1000 & 0.38 & 0.72 & 0.26 \\ 
  \addlinespace
  GSM8K    & Full data set     & 8737 & 0.52 & 0.82 & 0.26 \\
           & Test set of subsampled data set &1000 & 0.29 & 0.77 & 0.28 \\
  \addlinespace
  TriviaQA & Full data set     & 4950 & 0.74 & 0.87 & 0.11 \\
           & Test set of subsampled data set & 1000 & 0.44 & 0.79 & 0.16 \\
  \bottomrule
\end{tabular}
\label{tab:AppendixConsistencyFullLlama}
\end{table}

\end{document}